\documentclass[preprint,12pt，fleqn]{elsarticle}

\usepackage{natbib}
\usepackage{amssymb}
\usepackage{algpseudocode}
\usepackage{hyperref}
\usepackage{amsmath}
\usepackage{xcolor}
\usepackage{algorithm}
\usepackage{fancyhdr}
\usepackage{graphicx}
\usepackage{subfig}        
\usepackage{float}         
\usepackage{makecell}
\usepackage{multirow}  

\begin{document}

\begin{frontmatter}

\title{Multi-view Graph Convolutional Network with Fully Leveraging Consistency via Granular-ball-based Topology Construction, Feature Enhancement and Interactive Fusion}

\author[a]{Chengjie Cui}
\author[a]{Taihua Xu\corref{cor1}}
\author[b]{Shuyin Xia}
\author[b]{Qinghua Zhang}
\author[a]{Yun Cui}
\author[c]{Shiping Wang}
\cortext[cor1]{Corresponding author.\\
\hspace*{15pt}\textit{E-mail address:} 231210703107@stu.just.edu.cn (Chengjie Cui), xutaihua2019@just.edu.cn (Taihua Xu), xiasy@cqupt.edu.cn (Shuyin Xia), zhangqh@cqupt.edu.cn (Qinghua Zhang),  ycui@just.edu.cn (Yun Cui), shipingwangphd@163.com (Shiping Wang).}
\address[a]{ School of Computer, Jiangsu University of Science and Technology, Zhenjiang, 212100, China} 
\address[b]{ Chongqing Key Laboratory of Computational Intelligence, Chongqing University of Posts and Telecommunications, Chongqing, 400065, China} 
\address[c]{ College of Computer and Data Science, Fuzhou University, Fuzhou 350116, China}
\begin{abstract}
The effective utilization of consistency is crucial for multi-view learning. Graph convolutional networks (GCNs) leverage node connections to propagate information across the graph, facilitating the exploitation of consistency in multi-view data. However, most existing GCN-based multi-view methods suffer from several limitations. First, current approaches predominantly rely on \(k\)-Nearest Neighbors (\textit{k}NN) for topology construction, where the artificial selection of the \(k\) value significantly constrains the effective exploitation of inter-node consistency. Second, the inter-feature consistency within individual views is often overlooked, which adversely affects the quality of the final embedding representations. Moreover, these methods fail to fully utilize inter-view consistency as the fusion of embedded representations from multiple views is often implemented after the intra-view graph convolutional operation. Collectively, these issues limit the model's capacity to fully capture inter-node, inter-feature and inter-view consistency. To address these issues, this paper proposes the multi-view graph convolutional network with fully leveraging consistency via granular-ball-based topology construction, feature enhancement and interactive fusion (MGCN-FLC). MGCN-FLC can fully utilize three types of consistency via the following three modules to enhance learning ability:
(1) The topology construction module based on the granular ball algorithm, which clusters nodes into granular balls with high internal similarity to capture inter-node consistency; (2) The feature enhancement module that improves feature representations by capturing inter-feature consistency;
(3) The interactive fusion module that enables each view to deeply interact with all other views, thereby obtaining more comprehensive inter-view consistency.
Experimental results on nine datasets show that the proposed MGCN-FLC outperforms state-of-the-art  semi-supervised node classification methods.
\end{abstract}

\end{frontmatter}

\section{Introduction}
Multi-view data refers to data obtained from multiple perspectives, sources, or feature extraction methods that describe the same set of objects. While the feature sets of different views may vary, the objects across all views maintain a one-to-one correspondence. Multi-view learning can comprehensively utilize information from multiple views to achieve better representational capability, and has sparked extensive research across multiple fields including computer vision \cite{DBLP:conf/cvpr/FazlaliXRL22,DBLP:journals/inffus/LupionRQSO24,DBLP:journals/nn/WeiHWYZL24} and machine learning \cite{DBLP:journals/tmm/HuangZFW23,DBLP:journals/isci/ZhangDLZCW23,DBLP:conf/aaai/XuSGZ0G24}. The effective utilization of consistency is crucial for multi-view learning \cite{cheng2021multi}. By using the consistency in multi-view data \cite{DBLP:journals/nn/HuangZLYZZ25,DBLP:journals/isci/ZhongLCSW24}, multi-view learning can achieve high-quality embedding representations.  


Graph neural networks (GNNs) have demonstrated strong expressive power in various learning tasks \cite{DBLP:conf/nips/WuZF23,DBLP:journals/eswa/WangZYHCW22,wang2024graph,yu2024sdhgcn}, driving significant attention in the field of multi-view learning. In particular, graph convolutional networks (GCNs) \cite{DBLP:conf/iclr/KipfW17}, a widely used GNN model, extend convolutional operations to non-Euclidean graph data, enabling node features to interact through topology structure. GCNs have been applied successfully in downstream tasks such as node classification \cite{DBLP:conf/kdd/SongZK22,DBLP:conf/kdd/YueLCB22,DBLP:journals/tmm/LuWZCZW24} and link prediction \cite{DBLP:journals/tkdd/AtaFWSKL21,DBLP:journals/corr/abs-2206-04216}. Recently, numerous GCN-based multi-view learning approaches \cite{DBLP:journals/isci/ZhongLCSW24,DBLP:journals/inffus/ChenFYGPW23,DBLP:journals/eswa/PengDC25,DBLP:journals/kbs/WangLWGW25} have incorporated GCNs into multi-view data. These methods can effectively leverage the consistency \cite{DBLP:journals/nn/HuangZLYZZ25,DBLP:journals/isci/ZhongLCSW24} in multi-view data through message passing mechanisms on graphs, achieving superior learning capabilities \cite{DBLP:journals/eswa/PengDC25,DBLP:journals/eswa/WangYZXYX26}.

Since multi-view data does not provide the topology information, the construction of high-quality topology structures is a necessary step in GCN-based multi-view learning. Some studies \cite{DBLP:conf/aaai/LiLW20a,DBLP:journals/nn/ChenWCDW23} adopted \(k\)-Nearest Neighbors (\textit{k}NN) as the foundational algorithm for topology construction. Given that \textit{k}NN may introduce \textit{k}-value noise \cite{DBLP:journals/tcsv/WangCFWZ23,DBLP:conf/aaai/YangYPYF23,DBLP:journals/tmm/WuLLCBW23} into the constructed topology, some research \cite{DBLP:journals/isci/ZhongLCSW24,DBLP:journals/inffus/ChenFYGPW23,DBLP:journals/nn/ChenWCDW23} has made various improvements to the \textit{k}NN-based topology construction. However, a fixed global \textit{k} value is destined to not adapt to the data distribution of different views simultaneously. 
Then some latest studies \cite{DBLP:journals/eswa/PengDC25,DBLP:journals/eswa/WangYZXYX26,DBLP:journals/nn/DornaikaBCX25} moved away from the \textit{k}NN and instead focus on exploring novel methods for constructing topologies. Among them, GBCM-GCN \cite{DBLP:journals/eswa/WangYZXYX26} first introduces the unsupervised granular ball algorithm \cite{DBLP:journals/tnn/ChengLXWHZ24} to cluster the data into several granular balls (GBs), then the topology are constructed by performing node-level full connection between the GBs pairs according to the boundary distances. As is well known, the label consistency assumption \cite{DBLP:journals/tkde/BiDFWHZ24,DBLP:conf/www/DuSFMLHZ22} underpins the message passing mechanism in GNNs, and the quality of the learned embeddings strongly depends on how well the data satisfy this assumption. A commonly used metric to quantify this property is the homophily ratio \cite{DBLP:journals/tkde/BiDFWHZ24,DBLP:conf/www/DuSFMLHZ22}, defined as  the proportion of edges whose endpoints share the same label  relative to the total number of edges; this metric can also serves as an indicator of topology quality. Our analysis reveals that the homophily ratio of the GBCM-GCN \cite{DBLP:journals/eswa/WangYZXYX26} method is not very high, which inevitably impacts the quality of the constructed topology. Therefore, the first issue focused on in this paper is: how to construct the high-quality topologies with the high homophily ratio.

In recent years, leveraging feature interaction to enhance inter-feature consistency has emerged as a key research direction in GNNs \cite{DBLP:journals/tkde/ZhaoYLZGZ23,DBLP:journals/corr/abs-2003-02587}. DFI-GCN \cite{DBLP:journals/tkde/ZhaoYLZGZ23} employs  Newton's identities to extract high-order interactive features, while Cross-GCN \cite{DBLP:journals/corr/abs-2003-02587} introduces a "cross-feature graph convolution" operator to efficiently and explicitly capture interactive features of arbitrary orders. Although there is already research \cite{DBLP:journals/isci/ZhongLCSW24} successfully integrates feature interaction into multi-view data, it concentrates solely on inter-view feature interactions, overlooking intra-view feature interactions. This fails to fully leverage the inter-feature consistency within individual views, and degrades the quality of the final embeddings. Therefore, the second issue focused on in this paper is: how to effectively embed the inter-feature consistency within each view into the final representation.

Inter-view information interaction  directly affects the effectiveness of multi-view learning \cite{DBLP:journals/isci/ZhongLCSW24,xu2020deep}. Therefore, the utilization of inter-view consistency has always been a key research in multi-view learning \cite{DBLP:journals/inffus/ChenFYGPW23,DBLP:journals/eswa/PengDC25,DBLP:journals/eswa/WangYZXYX26,DBLP:journals/nn/ChenWCDW23}. In the GCN-based multi-view learning, the most traditional approach \cite{DBLP:journals/kbs/WangLWGW25,DBLP:journals/nn/WangHWLCZ24,DBLP:journals/air/WangLCWHZ25,DBLP:journals/eswa/WuCZWG25} to inter-view feature interaction is to directly concatenate feature representations from all views. However, this strategy neither enables deep interaction among views nor explicitly exploits inter-view consistency. Recently, GBCM-GCN \cite{DBLP:journals/eswa/WangYZXYX26} proposed constructing the layer-specific collaboration matrices using the view with the richest feature information and sharing them across convolutional layers of other views. Although this approach explicitly leverages inter-view consistency, the collaboration matrices are derived solely from information-richest view, which limits information exchange across arbitrary views. As a result, GBCM-GCN \cite{DBLP:journals/eswa/WangYZXYX26} does not fully utilize inter-view consistency. Therefore, the third issue focused on in this paper is: how to explicitly and fully leverage the inter-view consistency by the information interaction across arbitrary views. 

To address the above three issues, this paper proposes a multi-view graph convolutional network with fully leveraging consistency via granular-ball-based topology construction, feature enhancement and interactive fusion (MGCN-FLC). The MGCN-FLC consists of three modules: topology construction, feature enhancement and interactive fusion.

\textbf{Topology construction module.} The topology construction module employs an unsupervised granular ball clustering algorithm\cite{DBLP:journals/tnn/ChengLXWHZ24} to cluster nodes into granular balls (GBs). The nodes within the same GB demonstrate a high-level similarity, leading to strong categorical consistency. Moreover, nodes from two closely positioned GBs also display similarity to some extent. Based on GBs, this module performs node connections at two levels: intra-GB connections and inter-GB connections, which yields the high-quality topologies with the high homophily ratio.

\textbf{Feature Enhancement Module.} The feature enhancement module introduces feature interaction within each individual view. Specifically, for each view, the inter-feature similarity matrix is computed and then is multiplied by the original features matrix to generate similarity-based features matrix. To extract more comprehensive information, mixed pooling that combines average pooling and max pooling is applied to the stacked original and similarity-based feature matrices, producing an enhanced feature matrix. This enhanced feature matrix contains the results of intra-view feature interaction, effectively enhancing inter-feature consistency representation.

\textbf{Interactive Fusion Module.} The interactive fusion module is designed to deeply explore the inter-view interaction information to explicitly and fully exploit inter-view consistency.
Specifically, the interaction matrix between any two views is computed, and then the interaction matrices corresponding to each view are aggregated together, which can explicitly express the inter-view consistency. Furthermore, all views share a common weight matrix to generate the final feature representation, which can fully exploit the inter-view consistency.

The main contributions of this paper are summarized as follows:

\begin{itemize}
\item{An adaptive topology construction method is proposed, which constructs two-level node connections (intra-GB and inter-GB) via unsupervised GB clustering, yielding high-homophily topologies.}

\item{A feature enhancement method is designed, enhancing intra-view feature interaction and inter-feature consistency representation by computing the feature similarity matrix and applying mixed pooling.}

\item{An interactive fusion module is used to explicitly express and fully exploit the inter-view consistency.}

\item{In the downstream semi-supervised node classification task, experimental results demonstrate the effectiveness of the proposed MGCN-FLC.}
\end{itemize}

\section{Related work}
We will introduce the granular ball, graph convolutional  networks, and GCN-based multi-view learning.
\subsection{Granular balls}
Most GCN-based multi-view learning \cite{DBLP:journals/tcsv/WangCFWZ23, DBLP:conf/aaai/YangYPYF23} employed \(k\)-Nearest Neighbors (\textit{k}NN) for topology construction. However, the use of \textit{k}NN inevitably introduces noise associated with the selection of the  \(k\) value, which can degrade the quality of the constructed topology. To address this issue, Xia et al. \cite{DBLP:journals/isci/XiaLDWYL19} proposed a clustering algorithm that avoids \(k\)-value noise by partitioning data into granular balls (GBs). This partitioning is based on the premise that spatially proximate objects tend to exhibit similar distributions and are therefore likely to belong to the same category. This premise naturally leads to the need for evaluating GB's quality. In other words, an optimal GB should contain as many objects of the same category as possible. For this issue, the concept of "purity" was proposed  \cite{DBLP:journals/tnn/XiaDWGG24} and adopted as the termination criterion for GB partitioning. However, the computation of purity heavily relies on the availability of label information. To overcome this limitation, Cheng et al. \cite{DBLP:journals/tnn/ChengLXWHZ24} developed an unsupervised GB clustering algorithm that adopts \( \sqrt{N} \) as the termination criterion for GB partitioning, thereby eliminating the reliance on label information. In this work, we adopt this unsupervised GB clustering method to generate GBs, which avoids can avoid \(k\)-value noise and does not require label information. As a result, it provides reliable node category information for subsequent topology construction.

\subsection{Graph convolutional network}
Graph Convolutional Networks (GCNs) \cite{DBLP:conf/iclr/KipfW17} are an important branch of Graph Neural Networks (GNNs), which extended the convolution operation from regular Euclidean data to non-Euclidean graph-structured data. The core idea of GCNs is to use the topology to guide the propagation and aggregation of node features, enabling each node to gather information from its neighbors, thereby generating node embeddings that encapsulate rich contextual relationships. 

GCN employed the following propagation formula (Eq. \ref{deqn_ex56}), which incorporates topology information (i.e., the adjacency matrix) to propagate and aggregate node features:

\begin{equation}
\label{deqn_ex56}
\mathbf{H}^{(l+1)} = \sigma \left( \tilde{\mathbf{D}}^{- \frac{1}{2}} \tilde{\mathbf{A}} \tilde{\mathbf{D}}^{-\frac{1}{2}} \mathbf{H}^{(l)} \mathbf{W}^{(l+1)} \right),
\end{equation}
where \(H^{(l+1)}\) represents the embedding at the \((l+1)\)-th layer of the GCN. To retain the node's own information after feature information integration, GCN introduces self-loops to the adjacency matrix \(A\), resulting in \(\tilde{A}=A+I\). \(\tilde{D}\) represents the degree matrix of \(\tilde{A}\). \({H}^{(l)}\) is the embedding at the \(l\)-th layer. \(W^{(l+1)}\) represents the weight matrix at the \((l+1)\)-th layer and \(\sigma(\cdot)\) is the activation function.

Based on GCN, many extension algorithms \cite{DBLP:journals/tkde/ZhaoYLZGZ23,DBLP:journals/tetc/HeCGS24,DBLP:conf/kdd/0017ZB0SP20,DBLP:journals/isci/WangLCL21,DBLP:journals/nn/LiuGTQSX21} have been developed that demonstrate superior performance in semi-supervised node classification tasks. DFI-GCN \cite{DBLP:journals/tkde/ZhaoYLZGZ23} utilizes Newton's identity to extract across features of different orders from the original features and designs an attention mechanism to fuse these features, enhancing feature representation capability. HDGCN \cite{DBLP:journals/tetc/HeCGS24} introduces a dual-channel architecture that effectively captures high-order topological information while preserving original features, addressing the lack of robustness in single-channel GCNs when processing high-order information. AM-GCN \cite{DBLP:conf/kdd/0017ZB0SP20} adaptively fuses multi-channel convolution results via an attention mechanism, effectively mitigating insufficient integration of node feature and topology information in GCN. MOGCN \cite{DBLP:journals/isci/WangLCL21} constructs multiple learners using multi-order adjacency matrices and introduces an ensemble module to fuse the results of these learners, alleviating the over-smoothing issue. LPD-GCN \cite{DBLP:journals/nn/LiuGTQSX21} utilizes an encoder-decoder mechanism to supervise each convolutional layer, enabling better preservation of local features in the hidden representations of each layer. 

\begin{figure}[h]
\centering
\includegraphics[width=\textwidth]{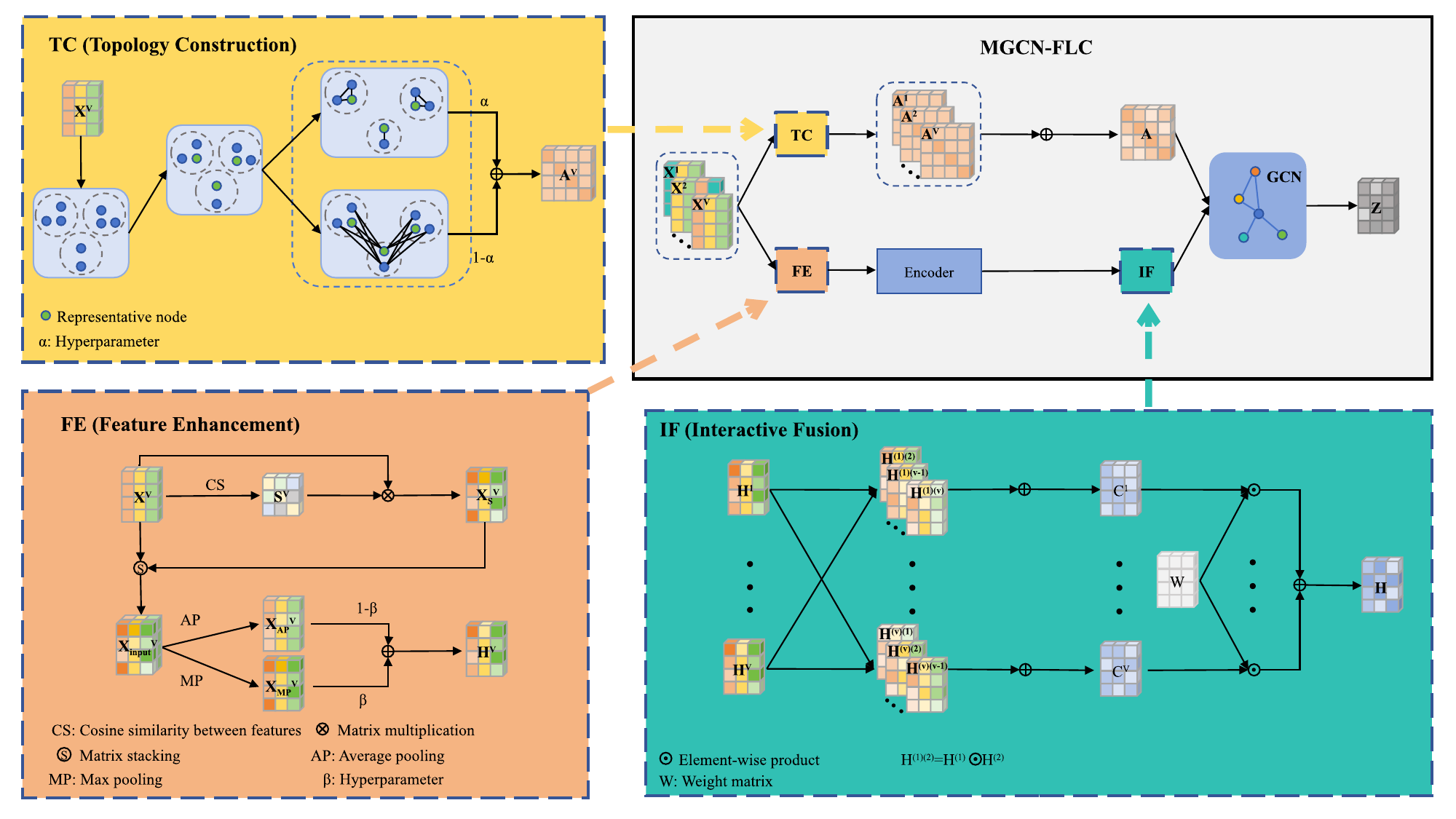}
\caption{MGCN-FLC consists of three modules: the topology construction module (TC), the feature enhancement module (FE) and the interactive fusion module (IF). Specifically, the TC module constructs the topology by establishing the inter-GB and intra-GB full connections between nodes, providing an effective pathway for information propagation in the GCN. The FE module applies mixed pooling to the stacking result of the similarity feature matrix and the original feature matrix, generating the enhanced feature representations. Subsequently, the enhanced feature representations generated by FE module from multiple views are encoding into the same dimension. Based on the encoder's output, the IF module computes and aggregates interactive features between two distinct views to ultimately generate node embeddings. Finally, the topology constructed by the TC module and the node embeddings generated by the IF module are jointly fed as input into a standard GCN for node prediction.}
\label{fig1}
\end{figure}

\subsection{GCN-based multi-view learning}
The core of multi-view learning is to capture the consistency and complementarity between multiple views, thereby enhancing the model's performance. A multi-view dataset can be represented as \( \mathcal{X}=\{ X^v \}_{v=1}^V\). For the \( v \)-th view, \( X^v = \left[ x_1, x_2, \dots, x_N \right]^T \in \mathbb{R}^{N \times d^v} \), where \( N \) is the number of nodes, and \( d^v \) is the number of features in the \( v \)-th view. From the perspective of feature, the multiple-view data can also be denoted as \( \mathcal{X}=\{ F^v \}_{v=1}^V\). For the \(v\)-th view, \(F^v=\left[ f_1, f_2, \dots, f_{d^{v}} \right] \in \mathbb{R}^{N \times d^v} \).

Researchers \cite{DBLP:journals/inffus/ChenFYGPW23,DBLP:journals/eswa/PengDC25,DBLP:journals/eswa/WangYZXYX26,DBLP:conf/aaai/LiLW20a,DBLP:journals/nn/ChenWCDW23,DBLP:journals/eswa/WuCZWG25} have been proposed various GCN-based methods for handling multi-view data.
LGCN-FF \cite{DBLP:journals/inffus/ChenFYGPW23} implements collaborative optimization between features and topologies in multi-view data by jointly training the feature fusion network and the adaptive topology fusion module. JFGCN \cite{DBLP:journals/nn/ChenWCDW23} learns consistent feature representations through the autoencoder, and introduces the adaptive fusion mechanism that combines \(k\)-Nearest Neighbors (\textit{k}NN) and \(k\)-Farthest Neighbors (\textit{k}FN), achieving collaborative optimization between features and topologies. Co-GCN \cite{DBLP:conf/aaai/LiLW20a} constructs \textit{k}NN-based topologies from the features of each view as a foundation, and provides each view with topology information derived from a weighted combination of all views' Laplacian matrices, thereby achieving implicit collaboration of inter-view topologies. 
CGCN-FMF \cite{DBLP:journals/eswa/PengDC25} employs both \textit{k}NN and limited label information to obtain the optimal topology for each view, dynamically fusing them via trainable weights \(\pi\), and introduces the 1D convolutional neural network to fuse multi-view features learned by sparse autoencoders into the unified representation.
MSGG \cite{DBLP:journals/eswa/WuCZWG25} constructs the topology by employing edge sampling and path sampling strategies to collect topology information, and utilizes maximum likelihood estimation to directly learn the integrated topology that captures cross-view consistency. Subsequently, the MSGG model uses MLP to extract multi-view features separately, and then concatenates these features to obtain the fused feature representation. GBCM-GCN \cite{DBLP:journals/eswa/WangYZXYX26} introduces an unsupervised GB algorithm \cite{DBLP:journals/tnn/ChengLXWHZ24} to construct topologies by fully connecting nodes between GB pairs based on boundary distances. In addition, it computes layer-specific collaborative matrices from the most feature-rich view and shares them across all views to enhance cross-view consistency. 

Although these methods have achieved promising performance in  multi-view learning, several limitations remain. These methods either fail to hinder the \(k\)-value noise from the \textit{k}NN applied in topology construction, or do not to fully exploit the inter-view consistency through the feature fusion strategies. In this paper, we continue to adopt the unsupervised GB algorithm used in previous work \cite{DBLP:journals/eswa/WangYZXYX26} to generate GBs, in which nodes belonging to the same GB naturally exhibit consistency. The difference lies in that the topology constructed in this paper includes both intra-GB and inter-GB connections, exhibiting a high homophily ratio. Furthermore, we introduce the feature enhancement module and the interactive fusion module. The feature enhancement module refines intra-view feature representation within each view by leveraging inter-feature interactions, while the interactive fusion module explicitly and fully explores the inter-view consistency.

\section{The proposed method}
In this section, we propose the MGCN-FLC model, which is designed to explore inter-node, inter-feature, and inter-view consistency in multi-view data. The model consists of three main modules: topology construction, feature enhancement, and interactive fusion. The overall framework of MGCN-FLC is illustrated in Fig.\ref{fig1}.

\subsection{Topology construction via GBs}
We employ an unsupervised GB clustering algorithm \cite{DBLP:journals/tnn/ChengLXWHZ24} to generate GBs. The specific process is as follows: all nodes in each view are  initially assigned to a single GB. To ensure high inter-node similarity within each GB while maintaining computational efficiency, the balance must be struck between the size and the quantity of GB. In \cite{DBLP:journals/tnn/ChengLXWHZ24}, \( \sqrt{N} \) is used as the termination condition for GB splitting.  GB is split if the number of nodes it contains exceeds \( \sqrt{N} \) ; otherwise, the splitting process terminates. This procedure is repeated iteratively until all GBs satisfy the termination condition. Upon completion, the resulting set of  GBs is denoted as \(GBs\)=\{ \(GB_1\), \(GB_2\), \ldots,\(GB_m\)\}. 
Algorithm \ref{alg:alg1} shows the detailed procedure for generating \(GBs\) in the \(v\)-th view.

For intra-GB topology construction, we apply a full connection operation within each \(GB_I\) (\(I\in [1,m]\)), resulting in intra-GB topology \(A_{\text{intra}}\). The element at the \(i\)-th row and \(j\)-th column of \(A_{\text{intra}}(i,j)\), denoted \(A_{\text{intra}}(i,j)\), is defined as follows:
\begin{equation}
\label{eq13}
A_{\text{intra}}(i, j)= 
\begin{cases} 
1 & \text{if } x_i, x_j \in GB_I, \\
0 & \text{otherwise},
\end{cases}
\end{equation}
where \(x_i\) and \(x_j\) are two different nodes. 

\begin{algorithm}[!t]
\caption{Generate-GBs}\label{alg:alg1}
\begin{algorithmic}
\State {\bf{Input:}} \( X^v = \left[ x_1, x_2, \dots, x_N \right]^T \in \mathbb{R}^{N \times d^v} \).
\State {\bf{Output:}} \(GBs\)=\{ \(GB_1\), \(GB_2\), \ldots, \(GB_m\)\}.
\State Initialize: \(GB=\left[ x_1, x_2, \dots, x_N \right]^T \in \mathbb{R}^{N \times d^v}\), \(GBs\)=\(\varnothing\).
\State Add \(GB\) to an empty queue \(Q\).
\State \textbf{While} \(Q\) is not empty \textbf{do} 
\State \hspace{0.2cm} Get the first \(GB\) from \(Q\) and delete it from \(Q\).
\State \hspace{0.2cm} \textbf{if} the \(GB\) contains more than \( \sqrt{N} \) nodes \textbf{then}
\State \hspace{0.7cm}Split the \(GB\) into two sub-GBs using the 2-means algorithm.
\State \hspace{0.7cm}Enqueue the two sub-GBs into \(Q\).
\State \hspace{0.2cm} \textbf{end if} the \(GB\) contains no more than \( \sqrt{N} \) nodes \textbf{then}
\State \hspace{0.7cm}Add the \(GB\)  to \(GBs\).
\State \hspace{0.2cm} \textbf{end} 
\State \textbf{End} 
\State \textbf{Return:} \text{\( GBs \)}
\end{algorithmic}
\label{alg1}
\end{algorithm}

To enable nodes to learn global information, the inter-GB connections also need to be established. We propose connecting the two most similar GBs. Specifically, for each GB, a representative node is selected, which is the node within the GB that minimizes the sum of Euclidean distances to all other nodes.
Formally, the representative node \(r_I\) of \(GB_I\) is chosen as follows:
\begin{equation}
\label{eq14}
r_I = \arg \min_{x_i \in GB_I} \sum_{x_j \in GB_I} Dist(x_i,x_j),
\end{equation}
where \(Dist(x_i,x_j)=\| x_i - x_j \|_2\). The Euclidiean distance between two representative nodes \(r_I\) and \(r_J\), \(Dist(r_I, r_J)=\| r_I - r_J \|_2\), is then used to measure the similarity between two GBs \(GB_I\) and \(GB_J\). We use full connection operation 
between \(GB_I\) and its nearest neighbor \(GB_J\) to construct the inter-GB topology \(A_{\text{inter}}\). The element at the \(i\)-th row and \(j\)-th column of \(A_{\text{inter}}(i,j)\), denoted \(A_{\text{inter}}(i,j)\) is defined as follows:

\begin{equation}
\label{eq16}
A_{\text{inter}}(i, j) = 
\begin{cases} 
1 & \text{if } x_i \in GB_I, x_j \in GB_J, \\
0 & \text{otherwise},
\end{cases}
\end{equation}
 where \(x_i\) and \(x_j\) are two different nodes. 
 
 For the \(v\)-th view, the topology \(A^v\) is constructed by fusing 
 \(A^v_{\text{intra}}\) and \(A^v_{\text{inter}}\). The fusion is controlled by the hyperparameter \(\alpha\), which is learnable from the data distribution through end-to-end training using a cross-entropy loss function, eliminating the need for manual tuning. The formula is as follows:

\begin{equation}
\label{eq17}
A^v = \alpha A^v_{\text{inter}} + (1 - \alpha) A^v_{\text{intra}}.
\end{equation}

The differences between views can lead to variations in the topologies. To obtain a unified topology \(A\) that aligns with all views, we introduce an adaptive weighting method to integrate the topologies from all views, as shown in Eq. \ref{eq18}.

\begin{equation}
\label{eq18}
A = \sum_{v=1}^{V} \pi^v A^v,
\end{equation}
where \(\pi^{v}\) is the weight of the \(v\)-th view, reflecting the importance of each view. 

To avoid any view from dominating the fusion of topologies, the weights of \(\pi^{v}\) are normalized as follows:

 \begin{equation}
	\label{eq19}
	\pi^v \leftarrow \frac{\exp(\pi^v)}{\sum\limits_{v=1}^{V} \exp(\pi^v)}.
\end{equation}

\subsection{Feature enhancement module}
To explore the inter-feature consistency within each view, the intra-view feature interaction is essential. 

Firstly, we design the similarity matrix \( S^{v} \in \mathbb{R}^{d^v \times d^v} \) for the \(v\)-th view by computing the similarities between all feature, characterizing the inter-feature consistency. The calculation formula is as follows:
\begin{equation}
	\label{eq3}
	S^{v} = \frac{(F^{v})^T F^{v}}{\| (F^{v})^T \|_2 \| (F^{v}) \|_2},
\end{equation}
where \( F^{v} \in \mathbb{R}^{N \times d^v} \) is the feature matrix of the $v$-th view, and \(\|\cdot\|_2\) represents \(L\)-2 norm.

To enable the original features to acquire inter-feature consistency information, the feature matrix \(X^{v}\) is transformed into the similarity feature matrix \(X^{v}_s\) via multiplication with \(S^{v}\):
\begin{equation}
	\label{eq4}
	X^{v}_s = X^{v}S^{v}.
\end{equation}

To obtain the comprehensive feature representation  \(H^v\) for the \(v\)-th view, we adopt the mixed pooling strategy. This strategy leverages max pooling to extract salient local details from the original feature matrix and average pooling to preserve global contextual information from the similarity feature matrix, thereby integrating both information sources. The formula is as follows:

\begin{equation}
	\label{eq5}
	H^v = \beta \text{MP}(X^v_{input}) + (1-\beta ) \text{AP}(X^v_{input}),
\end{equation}
where \(X^v_{input} \in \mathbb{R}^{N \times d^v \times 2} \) is the stacking of \(\mathbf{X}^v\) and \(\mathbf{X}_s^v\), and MP and AP represent max pooling and average pooling respectively. \(\beta\) is the hypeparameter. 

\subsection{Interactive fusion module}
 Most existing approaches do not explicitly or fully exploit inter-view consistency, which can result in suboptimal node embedding. In this work, we leverage inter-view consistency through interactive fusion to generate feature representations that yield more powerful node embeddings. Before performing the fusion, it is necessary to align feature dimensions across all views. To achieve this, we employ the sparse autoencoder \cite{DBLP:journals/inffus/ChenFYGPW23} to map features of different dimensions into the same dimension \(d\). The sparse autoencoder is formulated as follows:
\begin{equation}
	\label{eq7}
    O^{(l,v)} = \sigma \left( O^{(l-1,v)} W^{(l,v)}_{sa} + b^{(l,v)}_{sa} \right),
\end{equation}
where $O^{(l,v)} \in \mathbb{R}^{N \times d_l}$ represents the output of the \textit{l}-th layer in the sparse autoencoder for \(v\)-th view. \(d_l\) denotes the output feature dimensionality of the decoder in the sparse autoencoder. $W^{(l,v)}_{sa} \in \mathbb{R}^{d_{l-1} \times d_l}$, $b^{(l,v)}_{sa} \in \mathbb{R}^{d_l}$ and $\sigma$ represent the weights, the biases and the activation function, respectively. 

The loss function of  the sparse autoencoder for the \(v\)-th view is defined as follows:
\begin{equation}
	\label{eq8}
    \mathcal{L}^v_{sa} = \frac{1}{2} \left\| O^{(l,v)} - X^{v} \right\|_2^2 + \gamma D_{KL}(\rho \parallel \hat{\rho}),
\end{equation}
where \(\|\cdot\|_2\) represents \(L\)-2 norm, \(D_{KL}(\rho \parallel \hat{\rho})\) quantifies the divergence between the target sparsity \(\rho\) and the actual sparsity \(\hat{\rho}\) distributions, thereby guiding the sparse autoencoder to learn meaningful features. The parameter \(\gamma\) controls the strength of the sparsity.

The sparse autoencoder, consisting of the total of \(L\) layers, has two main components: the encoder and the decoder. Because the encoder produces dimensionally aligned feature representations, we adopt its output as the feature representation for the $v$-th view: \(H^{v}\)=\(O^{(\frac{L}{2},v)}\). 

To explicitly and fully leverage inter-view consistency, we explore the interactive information  between any two different views. For the node \(x_i\), the node-level interactive information \(c^{(v, v')}_i \in R^{N \times d} \) is calculated by using the feature information of \(x_i\) from \(v\)-th and \(v'\)-th views, i.e. \(H^{v}_i\) and \(H^{v'}_i\). Specifically, 
\begin{equation}
	\label{eq9}
	c^{(v, v')}_i = H^{v}_i \odot  H^{v'}_i,
\end{equation}
where \(\odot\) is the element-wise product. The interactive information involving \(N\) nodes is as follows:
\begin{equation}
	\label{eq10}
	c^{(v,v')} = \left[ c^{(v,v')}_1; c^{(v,v')}_2; \dots; c^{(v,v')}_N \right].
\end{equation}

The feature representation of the \(v\)-th view, \(C^{v}\in \mathbb{R}^{N \times d}\), is generated by aggregating the interactive information between the \(v\)-th view and all other views.
\begin{equation}
	\label{eq11}
 C^{v} = \sum_{v'=1 }^{V}  c^{(v, v')} - c^{(v, v)}.
\end{equation}

Finally, the output of the interactive fusion module, \(H \in \mathbb{R}^{N \times d}\), is computed by fusing the feature representations of all views using shared weights matrix which is trained end-to-end with the cross-entropy loss function:
\begin{equation}
	\label{eq12}
	H = \sum_{v=1}^{V} \sigma \left( C^{v}\odot W_{\text{share}} + b_{\text{share}} \right),
\end{equation}
where \(W_{share} \in \mathbb{R}^{N \times d} \) and \(b_{share} \in \mathbb{R}^{N\times d} \) are the share weights and biases, respectively.

\subsection{Semi-supervised node classification}
For the downstream semi-supervised node classification task, the constructed topology \(A\) (Eq. \ref{eq18}) and the fused feature representations  \(H\) (Eq. \ref{eq12}) are input into the GCN for information propagation. The propagation mechanism of features across GCN layers is defined by the forward propagation formula, which iteratively updates node features based on the topology:
 \begin{equation}
	\label{eq20}
	H^{l} = \sigma \left( A H^{l-1} W^{l} \right),
\end{equation}
where \(H^{l-1}\) is the updated feature representation from the \((l-1)\)-th layer, \(W^l\) is the trainable weight matrix of the \(l\)-th layer and \(\sigma \) is the non-linear activation function. In particular,  \(H^{1} = \sigma( AH^0W^{1} )\), with \(H^0=H\).

The traditional shallow GCN consists of two layers, then the node embedding \(Z\) is shown as follows:

 \begin{equation}
	\label{eq21}
    Z = \text{softmax}\left( A \sigma \left( A H W^{1} \right) W^{2} \right),
\end{equation}
where $softmax(\cdot)$ computes the probability distribution over node categories. 

For semi-supervised node classification tasks, the cross-entropy loss function is used to measure the discrepancy between the model's predicted probability distribution and the actual labels. Minimizing this loss ensures optimal classification performance. The loss function \(\mathcal{L}_{gcn}\) is shown as follows:
 \begin{equation}
	\label{eq22}
    \mathcal{L}_{gcn} = - \sum_{i \in X_L} \sum_{c=1}^{C} Y_{ic} \log(Z_{ic}),
\end{equation}
where \(X_L\) represents the set of labeled samples, \(C\) is the number of categories, \(Y_{ic}\) is the true label and \(Z_{ic}\)  represents the predicted probability for node $i$ belonging to class $c$.

Algorithm \ref{alg:alg2} shows the detailed procedure of MGCN-FLC.

\subsection{Complexity analysis}

This section analyzes the computational complexity of
the proposed MGCN-FLC model. 

For the $v$-th view, the topology construction module first employs the unsupervised GB algorithm \cite{DBLP:journals/tnn/ChengLXWHZ24} to generate GBs. Its time complexity is $O(N \log m)$ ($m \ll N$) in which $N$ and $m$ represent the number of nodes and GBs, respectively. Then, the module establishes node connections based on the GBs. On average, each GB contains approximately $\frac{N}{m}$ nodes, so the time complexity for  establishing intra-GB or inter-GB connections is $O(\frac{N^2}{m})$. For establishing inter-GB connections, representative nodes must be selected from each GB and the distances between each pair of GBs computed. Selecting representative nodes has the time complexity of $O(\frac{N^2}{m})$, while computing distances between GB pairs requires $O(m^2)$. In summary, the total time complexity of the topology construction module is $O(N \log m + \frac{N^2}{m} + m^2)$.

\begin{algorithm}[!t]
\caption{MGCN-FLC}\label{alg:alg2}
\begin{algorithmic}
\State {\bf{Input:}} \( X = \{ X_1, X_2, \dots, X_v \} \).
\State {\bf{Output:}} Node embedding  \( Z \).
\State \textbf{For \( v = 1 \) $\rightarrow$  \( V \) do}
\State \hspace{0.2cm}Generate \(GBs\)=\{ \(GB_1\), \(GB_2\), \ldots,\(GB_m\)\} by the Algorithm \ref{alg:alg1}.
\State \hspace{0.2cm}Construct the intra-GB topology \(A_{\text{intra}}^v\) by Eq. \ref{eq13}.
\State \hspace{0.2cm}Construct the inter-GB topology \(A_{\text{inter}}^v\) by Eq. \ref{eq16}.
\State \hspace{0.2cm}Generate \(A^v\) through  \(A_{\text{intra}}^v\) and \(A_{\text{inter}}^v\) by Eq. \ref{eq17}.
\State \hspace{0.2cm}Generate the feature representation \(H^v\) based on \(X^{v}_s\) by Eq. \ref{eq5}.
\State \textbf{End for}
\State Compute the unified topology A by Eq. \ref{eq18}.
\State \textbf{While} not converging \textbf{do}
\State \hspace{0.2cm}\textbf{for \( v = 1 \) $\rightarrow$ \( V \) do}
\State \hspace{0.7cm}Compute \( O^{(L/2,v)} \) for the sparse autoencoder by Eq. \ref{eq7}.
\State \hspace{0.7cm}Calculate the feature representation \(C^{v}\) by Eq. \ref{eq11}.
\State \hspace{0.2cm}\textbf{end for}
\State \hspace{0.2cm}Compute the output \(H\) of interactive fusion module by Eq. \ref{eq12}.
\State \hspace{0.2cm}Compute \(H^l\) of the GCN by Eq. \ref{eq20}.
\State \textbf{End while}
\State {\bf{Return:}} Node embedding \text{\( Z \)}.
\end{algorithmic}
\label{alg2}
\end{algorithm}

For the $v$-th view, the feature enhancement module first computes the similarity matrix with complexity $O(N(d^v)^2)$, and then multiplies the original feature matrix with the similarity matrix, also $O(N (d^v)^2)$. Stacking the original and similarity feature matrices and applying mixed pooling to produce the feature representation $H^v$ has complexity $O(N d^v)$. Thus, the total complexity of the feature enhancement module is $O(N(d^v) + N(d^v)^2)$.

For the $v$-th view, the interactive fusion module first aligns feature dimensions using the sparse autoencoder, with time complexity  $O(Nd^vd^1 + Nd^1 d)$, where $d^1$ and $d$ represent the hidden layer dimensions of the sparse autoencoder. Computing the interactive information between the $v$-th view and all other views is $O((V-1)  N  d)$, and aggregating the interactive information also requires $O((V-1) N d)$. Therefore, the total complexity of the interactive fusion module is $O(N  d^v \ d^1 + N  d^1  d + (V-1)  N d)$.

For the $v$-th view, these three modules have the total time complexity of $
O(N  d^v  d^1 + N  d^1  d + (V-1) N  d + N(d^v) + N(d^v)^2 + N \log m + \frac{N^2}{m} + m^2)$.
For a multi-view dataset with $V$ views, assuming $D = \sum_{v=1}^{V} d^v$, then the total time complexity of the three modules will be $O( N  D  d^1 + N V d^1  d + N V (V-1) d + N  D + N  D^2 + N V \log m + V \frac{N^2}{m} + V m^2)$.

In addition, thereafter the graph convolution operation is adopted by MGCN-FLC, which requires the time complexity of $O(N^2 d)$. Thus, the  time complexity of MGCN-FLC is:
$O(N V (D  d^1 + d^1 d + (V-1) d + D + D^2 + \log m + \frac{N}{m}) + V  m^2 + N^2 d)$.
Generally, there are $N \gg V$, $N \gg m$ and $D \gg {d^1, d}$, then the time complexity of MGCN-FLC can be simplified to: $O(N D^2 + N^2)$.

\section{Experiment}
In this section, we conduct extensive experiments on semi-supervised node classification tasks, including comparative experiments, parametric sensitivity analysis, ablation experiments, experiments with varying label rates, visualization analysis, convergence analysis, runtime comparison, and experimental discussion.

\subsection{Datasets}
\begin{table*}[ht]
\caption{A brief description of multi-view datasets}\label{tab:table1}
\centering
\fontsize{6}{8}\selectfont
\setlength{\tabcolsep}{5pt}
\begin{tabular}{l |l l l l l}
\noalign{\hrule height 1pt}
Datasets &  \# samples &  \# views &  \# features &  \# classes &  \# data types\\
\hline
BBCnews & 685 & 4 & 4,659/4,633/4,665/4,684 & 5 & Textual documents\\
Caltech101-7 & 1474 & 6 & 48/40/254/1984/512/928 & 7 & Object images\\
MNIST & 10000 & 3 & 30/9/9 & 10 & Digit images\\
NUS-WIDE & 2400 & 6 & 64/73/128/144/225/500 & 8 & Object images\\
WebKB & 203 & 3 & 1,703/230/230 & 4 & Textual documents\\
BBCsports & 544 & 2 & 3183/3203 & 5 & Textual documents\\
NGs & 500 & 3 & 2000/2000/2000 & 5 & Textual documents\\
ProteinFold & 694 & 12 & 27/27/27/.../27 & 27 & Protein documents\\
Reuters & 1500 & 5 & 21531/24893/34279/.../11547 & 6 & Textual documents\\

\noalign{\hrule height 1pt}
\end{tabular}
\label{table 1}
\end{table*}
Table \ref{table 1}. presents detailed information about multi-view datasets used in our experiments, including the number of samples, the number of views, the number of features of each view, the number of categories, and the data types.

\subsection{Algorithm and parameter setup}
The proposed MGCN-FLC is compared with nine algorithms, including two classical multi-view learning algorithms (MLAN \cite{DBLP:conf/aaai/NieCL17} and MVAR \cite{DBLP:journals/tip/TaoHNZY17}), five GCN-based multi-view algorithms focusing on topology construction (Co-GCN \cite{DBLP:conf/aaai/LiLW20a}, DSRL \cite{DBLP:journals/pami/WangCDL22}, LGCN-FF \cite{DBLP:journals/inffus/ChenFYGPW23}, JFGCN \cite{DBLP:journals/nn/ChenWCDW23}, and HGCN-MVSC \cite{DBLP:journals/nn/WangHWLCZ24}), and two GCN-based multi-view algorithms simultaneously leverage topology construction and cross-view interactive fusion (MvRL-DP \cite{DBLP:journals/kbs/WangLWGW25} and GBCM-GCN \cite{DBLP:journals/eswa/WangYZXYX26}).

The dataset is split into training, validation, and testing sets with ratios of 10\%, 10\%, and 80\% of the total dataset size, respectively. The parameters of the nine baseline are set according to the configurations reported in their original papers. For the proposed MGCN-FLC model, we employ a 2-layer graph convolutional architecture with the dropout rate of 0.5 and the learning rate of 0.01. The ReLU and Softmax functions are used as activation functions in the hidden and output layers, respectively. The sparse autoencoder is structured with layer sizes \([1024, 256]\), trained with a learning rate of 0.001 and weight decay of 0.01. The proposed MGCN-FLC model is implemented
in PyTorch platform and executed on the machine equipped with the Intel i9-12400F
CPU, Nvidia RTX 4070 GPU and 12 GB RAM.
\subsection{Classification accuracy and F1-score}
\renewcommand{\arraystretch}{1.4}
\begin{table*}[ht]
\caption{The classification performance of all compared semi-supervised classification methods uses 10\% labeled samples as supervision, where the best performance is highlighted in bold and the second-best results are underlined.}
    \centering
    {\fontsize{6}{7}\selectfont
    \setlength{\tabcolsep}{0.8pt} 
    \begin{tabular}{l|*{9}{c}}
\noalign{\hrule height 1pt}
Methods/Datasets &  BBCnews &  Caltech101-7 &  MNIST &  NUS-WIDE &  WebKB  &  BBCSports & NGs & ProteinFold & Reuters\\
\noalign{\hrule height 1pt}
\hline
\makecell{\textbf{MLAN} \\ (2017)} & \begin{tabular}{c}74.1 \\ 73.5\end{tabular} & \begin{tabular}{c}76.3 \\ 75.6\end{tabular} & \begin{tabular}{c}87.7 \\ 86.1\end{tabular} & \begin{tabular}{c}34.1 \\ 33.2\end{tabular} & \begin{tabular}{c}73.9 \\ 41.2\end{tabular} & \begin{tabular}{c}80.5 \\ 80.1\end{tabular} & \begin{tabular}{c}85.5 \\ 83.3\end{tabular}& \begin{tabular}{c}35.6\\ 31.2\end{tabular}& \begin{tabular}{c}64.6 \\ 63.2\end{tabular}\\
\hline
\makecell{\textbf{MVAR}\\ (2017)} & \begin{tabular}{c}75.3 \\ 76.6\end{tabular} & \begin{tabular}{c}79.1 \\ 78.5\end{tabular} & \begin{tabular}{c}84.9 \\ 81.7\end{tabular} & \begin{tabular}{c}29.3 \\ 28.5\end{tabular} & \begin{tabular}{c}76.9 \\ 46.3\end{tabular} & \begin{tabular}{c}82.1 \\ 82.8\end{tabular} & \begin{tabular}{c}86.6 \\ 85.7\end{tabular}& \begin{tabular}{c}39.6 \\ 35.1\end{tabular}& \begin{tabular}{c}83.5\\ 81.8\end{tabular}\\
\hline
\makecell{\textbf{Co-GCN}\\ (2020)} & \begin{tabular}{c}81.9 \\ 78.1\end{tabular} & \begin{tabular}{c}83.1 \\ 80.2\end{tabular} & \begin{tabular}{c}90.2 \\ 90.1\end{tabular} & \begin{tabular}{c}27.5 \\ 28.1\end{tabular} & \begin{tabular}{c}72.6 \\ 39.3\end{tabular} & \begin{tabular}{c}86.3 \\ 84.9\end{tabular} & \begin{tabular}{c}84.2 \\ 83.6\end{tabular}& \begin{tabular}{c}41.2\\ 37.8\end{tabular}& \begin{tabular}{c}78.4\\ 73.7\end{tabular}\\
\hline
\makecell{\textbf{DSRL}\\ (2022)} & \begin{tabular}{c}84.5 \\ 83.1\end{tabular} & \begin{tabular}{c}85.3 \\ 84.6\end{tabular} & \begin{tabular}{c}88.9 \\ 87.1\end{tabular} & \begin{tabular}{c}43.6 \\ 43.2\end{tabular} & \begin{tabular}{c}82.5 \\ 58.2\end{tabular} & \begin{tabular}{c}90.9 \\ 93.7\end{tabular} & \begin{tabular}{c}88.7 \\ 84.5\end{tabular}& \begin{tabular}{c}46.5\\ 39.5\end{tabular}& \begin{tabular}{c}36.5\\ 25.3\end{tabular}\\
\hline
\makecell{\textbf{LGCN-FF}\\ (2023)} & \begin{tabular}{c}86.1 \\ 85.8\end{tabular} & \begin{tabular}{c}86.5 \\ 85.5\end{tabular} & \begin{tabular}{c}88.8 \\ 86.4\end{tabular} & \begin{tabular}{c}32.4 \\ 28.6\end{tabular} & \begin{tabular}{c}83.1 \\ 52.6\end{tabular} & \begin{tabular}{c}94.7 \\ 94.7\end{tabular} & \begin{tabular}{c}92.4 \\ 92.4\end{tabular}& \begin{tabular}{c}41.3\\ 35.0\end{tabular}& \begin{tabular}{c}64.9\\ 60.9\end{tabular}\\
\hline
\makecell{\textbf{JFGCN}\\ (2023)} & \begin{tabular}{c}85.6 \\ 85.4\end{tabular} & \begin{tabular}{c}92.2 \\ 86.3\end{tabular} & \begin{tabular}{c}91.5 \\ 90.5\end{tabular} & \begin{tabular}{c}48.4 \\ 45.5\end{tabular} & \begin{tabular}{c}80.1 \\ 58.3\end{tabular} & \begin{tabular}{c}92.9 \\ 92.8\end{tabular} & \begin{tabular}{c}90.2 \\ 90.5\end{tabular}& \begin{tabular}{c}38.7\\ 29.7\end{tabular}& \begin{tabular}{c}57.9\\ 39.7\end{tabular}\\
\hline
\makecell{\textbf{HGCN-MVSC}\\(2024)} & \begin{tabular}{c}90.7 \\ 90.2\end{tabular} & \begin{tabular}{c}92.5 \\ 91.2\end{tabular} & \begin{tabular}{c}92.5 \\ 92.2\end{tabular} & \begin{tabular}{c}49.2 \\ 46.3\end{tabular} & \begin{tabular}{c}82.3 \\ 59.5\end{tabular} & \begin{tabular}{c}96.9 \\ 97.0\end{tabular} & \begin{tabular}{c}93.1 \\ 92.8\end{tabular}& \begin{tabular}{c}51.2\\ 47.6\end{tabular}& \begin{tabular}{c}77.4\\ 73.7\end{tabular}\\
\hline
\makecell{\textbf{MVRL-DP}\\ (2025)} & \begin{tabular}{c}94.2 \\ 94.3\end{tabular} & \begin{tabular}{c}93.1 \\ {\bf93.0}\end{tabular} & \begin{tabular}{c}94.1 \\ 94.1\end{tabular} & \begin{tabular}{c}54.3 \\ 52.1\end{tabular} & \begin{tabular}{c}83.8 \\ \underline{62.2}\end{tabular} & \begin{tabular}{c}\underline{97.1} \\ \underline{97.1}\end{tabular} & \begin{tabular}{c}93.5 \\ 83.1\end{tabular}& \begin{tabular}{c}54.0\\ 47.8\end{tabular}& \begin{tabular}{c}64.4\\ 55.4\end{tabular}\\
\hline
\makecell{\textbf{GBCM-GCN}\\(2026)} & \begin{tabular}{c}\underline{98.0} \\ \underline{97.1}\end{tabular} & \begin{tabular}{c}\underline{94.2} \\ 87.3\end{tabular} & \begin{tabular}{c}{\bf98.0} \\ {\bf98.1}\end{tabular} & \begin{tabular}{c}\underline{96.5} \\ \underline{96.5}\end{tabular} & \begin{tabular}{c}\underline{85.3} \\ 55.2\end{tabular} & \begin{tabular}{c}96.7 \\ 95.2\end{tabular} & \begin{tabular}{c}\underline{94.5} \\ \underline{94.0}\end{tabular}& \begin{tabular}{c}\underline{58.3} \\ \underline{48.0}\end{tabular}& \begin{tabular}{c}\underline{97.6} \\ \underline{96.5}\end{tabular}\\
\hline
\noalign{\hrule height 1pt}
\makecell{\textbf{MGCN-FLC}} & \begin{tabular}{c}{\bf 98.4} \\ {\bf 97.9}\end{tabular} & \begin{tabular}{c}{\bf 97.5} \\ \underline{91.3}\end{tabular} & \begin{tabular}{c}\underline{96.4} \\ \underline{95.8}\end{tabular} & \begin{tabular}{c}{\bf 99.0} \\ {\bf 99.0}\end{tabular} & \begin{tabular}{c}{\bf91.8} \\ {\bf67.2}\end{tabular} & \begin{tabular}{c}{\bf97.4} \\ {\bf97.3}\end{tabular} & \begin{tabular}{c}{\bf97.8} \\ {\bf97.8}\end{tabular}& \begin{tabular}{c}{\bf60.0} \\ {\bf48.3}\end{tabular}& \begin{tabular}{c}{\bf 99.1} \\ {\bf98.9}\end{tabular}\\
\noalign{\hrule height 1pt}
    \end{tabular}
\label{table 2}}
\end{table*}
The accuracy (ACC) and F1-score of all algorithms on the nine datasets are presented in Table \ref{table 2}. On the Caltech101-7, NUS-WIDE, WebKB, NGs, ProteinFold, and Reuters datasets, MGCN-FLC outperforms the second-best method GBCM-GCN \cite{DBLP:journals/eswa/WangYZXYX26}, improving the ACC values by 3.3\%, 2.5\%, 6.5\%, 3.3\%, 1.7\%, and 1.5\%, respectively. 
These improvements are attributed to that
the proposed MGCN-FLC simultaneously utilizes both intra-ball and inter-ball connections, whereas GBCM-GCN \cite{DBLP:journals/eswa/WangYZXYX26} focuses only on the inter-ball connections, which facilitates feature propagation, thereby generating more expressive node embeddings.

\begin{figure*}[!t]
\centering
\subfloat[\footnotesize BBCnews]{\includegraphics[width=1.55in, trim=0cm 0cm 0.2cm 0.2cm, clip]{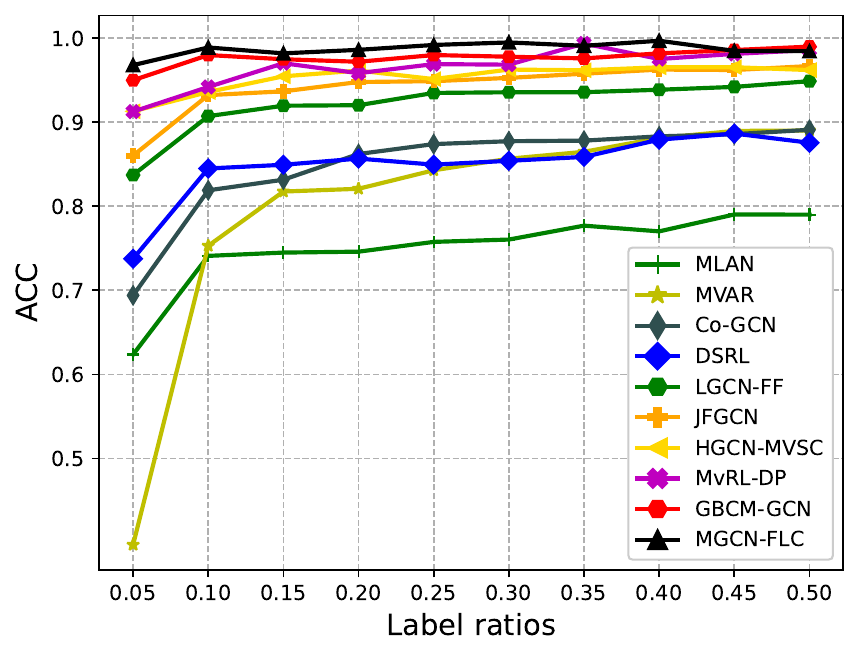}%
\label{fig_1_case}}
\vspace{-0.6cm}
\subfloat[\footnotesize Caltech101-7]{\includegraphics[width=1.55in, trim=0cm 0cm 0.2cm 0.2cm, clip]{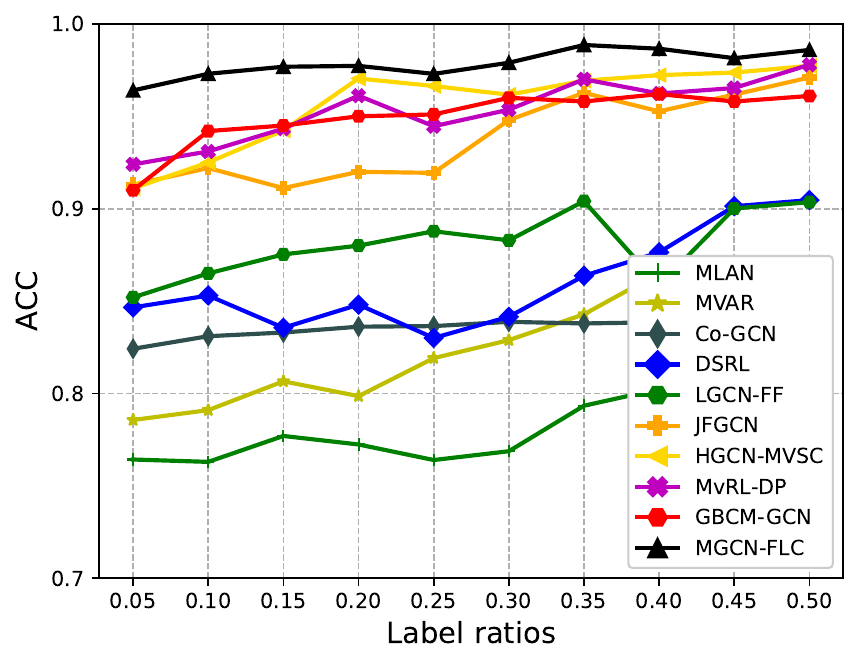}%
\label{fig_2_case}}
\vspace{-0.6cm}
\subfloat[\footnotesize MNIST]{\includegraphics[width=1.55in, trim=0cm 0cm 0.2cm 0.2cm, clip]{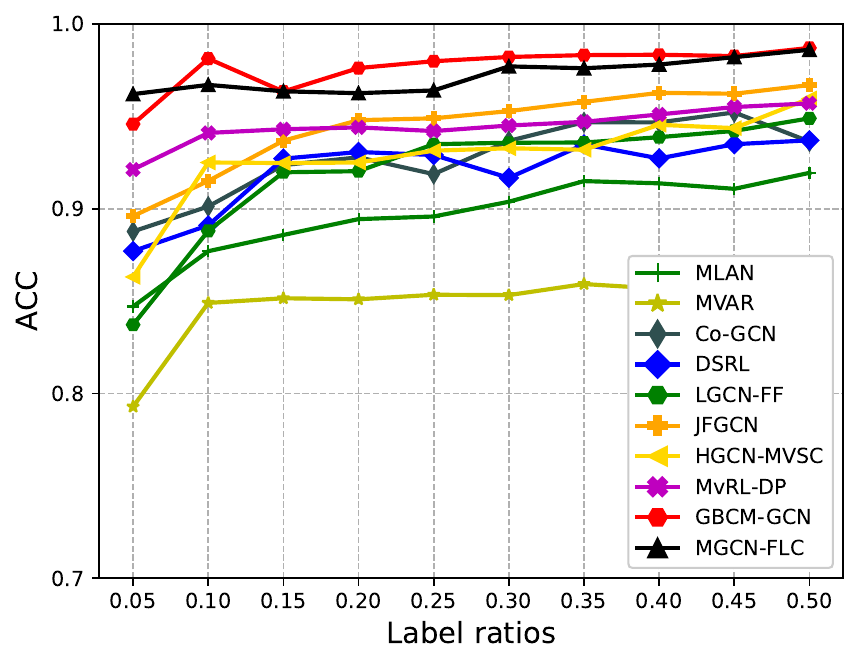}%
\label{fig_3_case}}
\vspace{1.1cm}
\subfloat[\footnotesize NUS-WIDE]{\includegraphics[width=1.55in, trim=0cm 0cm 0.2cm 0.2cm, clip]{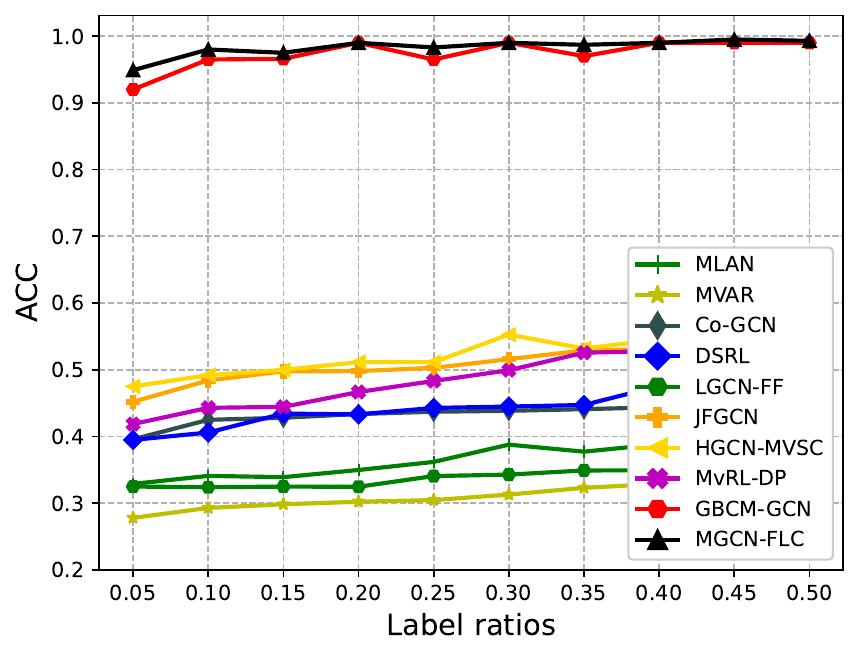}%
\label{fig_4_case}}
\vspace{-0.6cm}
\subfloat[\footnotesize WebKB]{\includegraphics[width=1.55in, trim=0cm 0cm 0.2cm 0.2cm, clip]{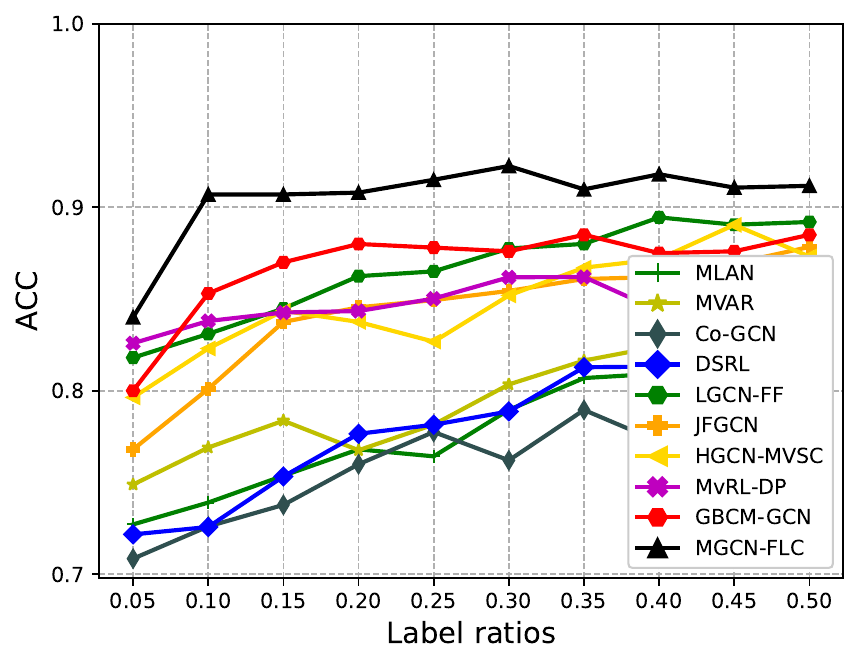}%
\label{fig_5_case}}
\vspace{-0.6cm}
\subfloat[\footnotesize BBCsports]{\includegraphics[width=1.55in, trim=0cm 0cm 0.2cm 0.2cm, clip]{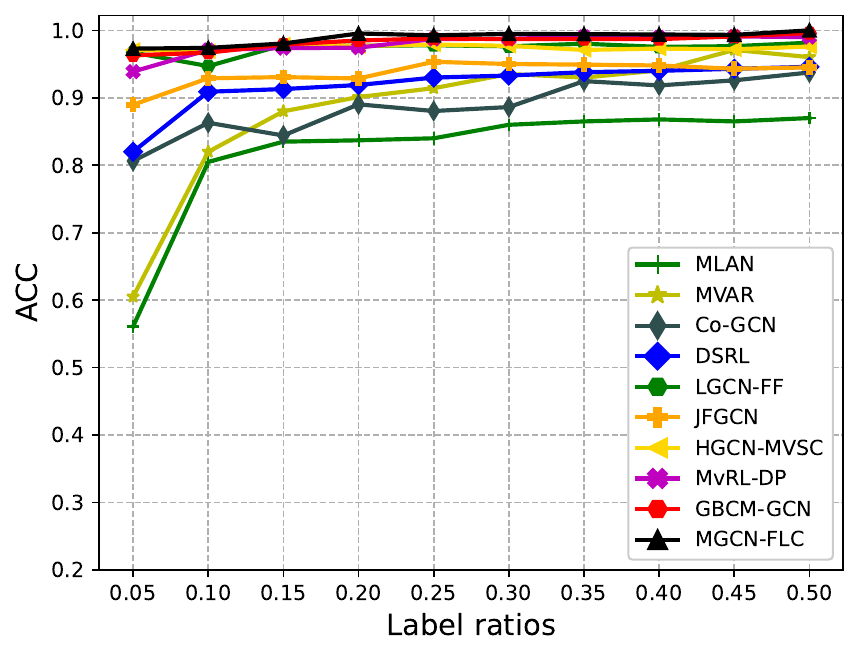}%
\label{fig_6_case}}
\vspace{1.2cm}
\subfloat[\footnotesize NGs]{\includegraphics[width=1.55in, trim=0cm 0cm 0.2cm 0.2cm, clip]{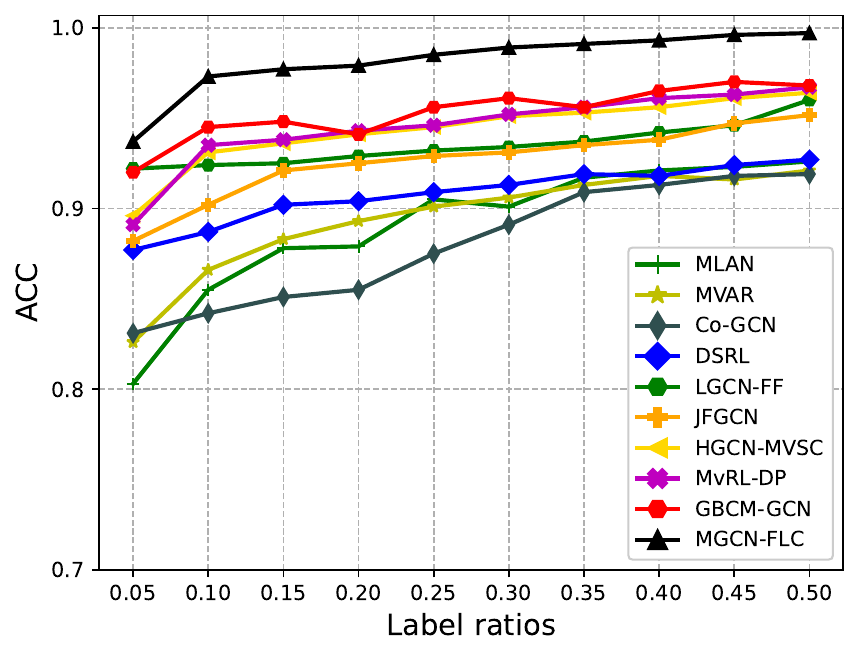}%
\label{fig_7_case}}
\vspace{-0.6cm}
\subfloat[\footnotesize ProteinFold]{\includegraphics[width=1.55in, trim=0cm 0cm 0.2cm 0.2cm, clip]{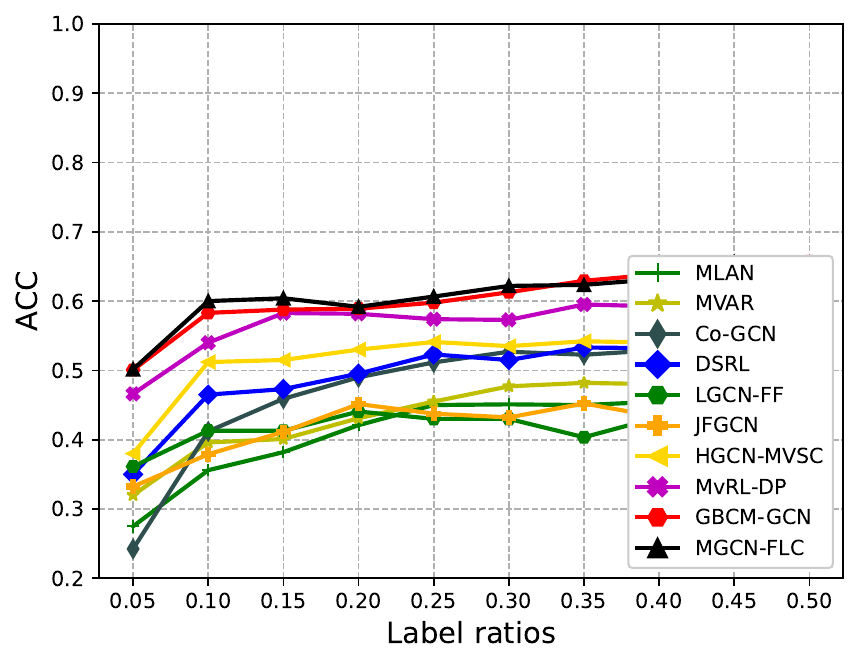}%
\label{fig_8_case}}
\vspace{-0.6cm}
\subfloat[\footnotesize Reuters]{\includegraphics[width=1.55in, trim=0cm 0cm 0.2cm 0.2cm, clip]{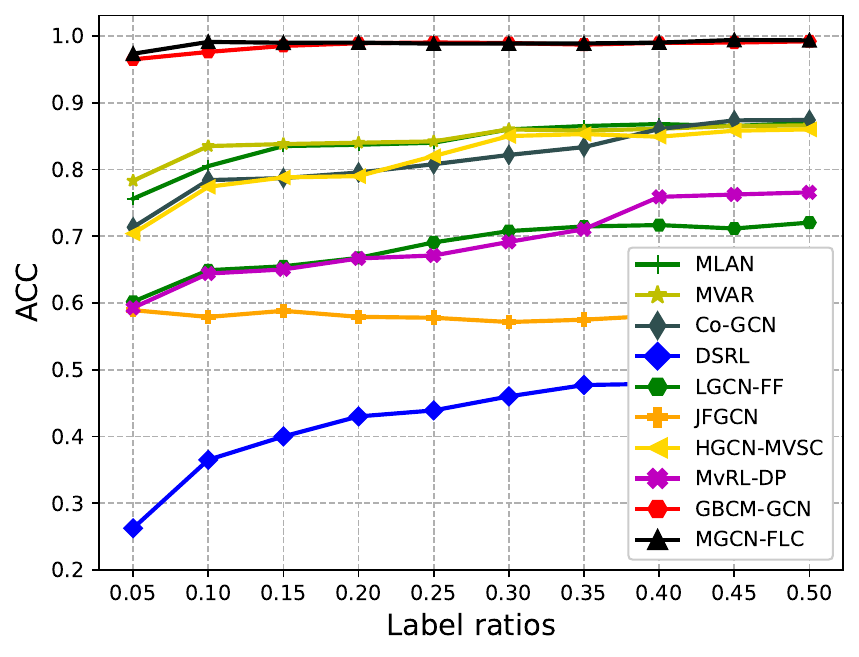}%
\label{fig_9_case}}
\vspace{1.2cm}
\caption{\small ACC of all methods as the ratio of labeled data ranges in \{0.05, 0.10, \dots , 0.5\} on the nine datasets.}
\label{fig_sim1}
\end{figure*}
\begin{figure*}[!t]
\centering
\subfloat[\footnotesize BBCnews]{\includegraphics[width=1.55in, trim=0cm 0cm 0.2cm 0.2cm, clip]{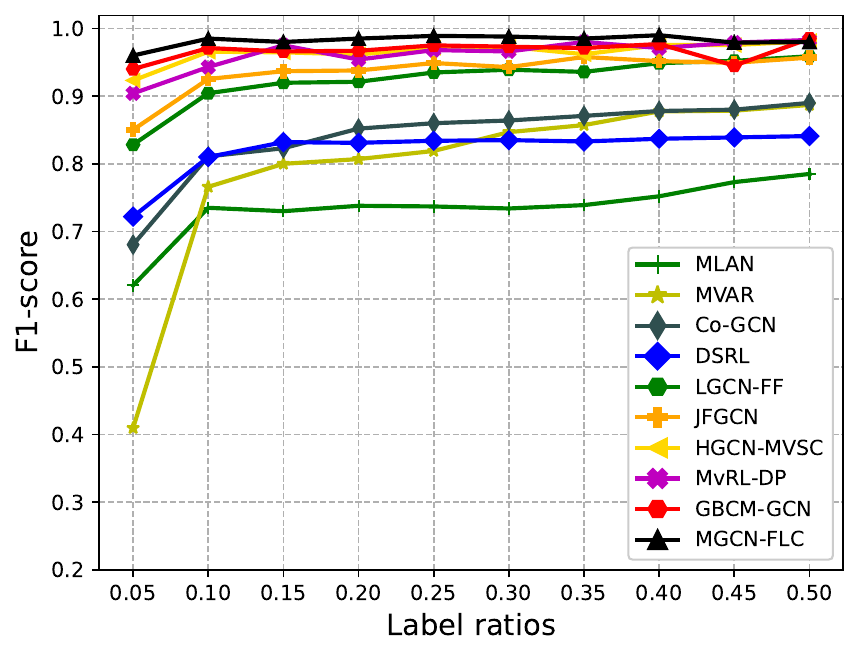}%
\label{fig_10_case}}
\vspace{-0.6cm}
\subfloat[\footnotesize Caltech101-7]{\includegraphics[width=1.55in, trim=0cm 0cm 0.2cm 0.2cm, clip]{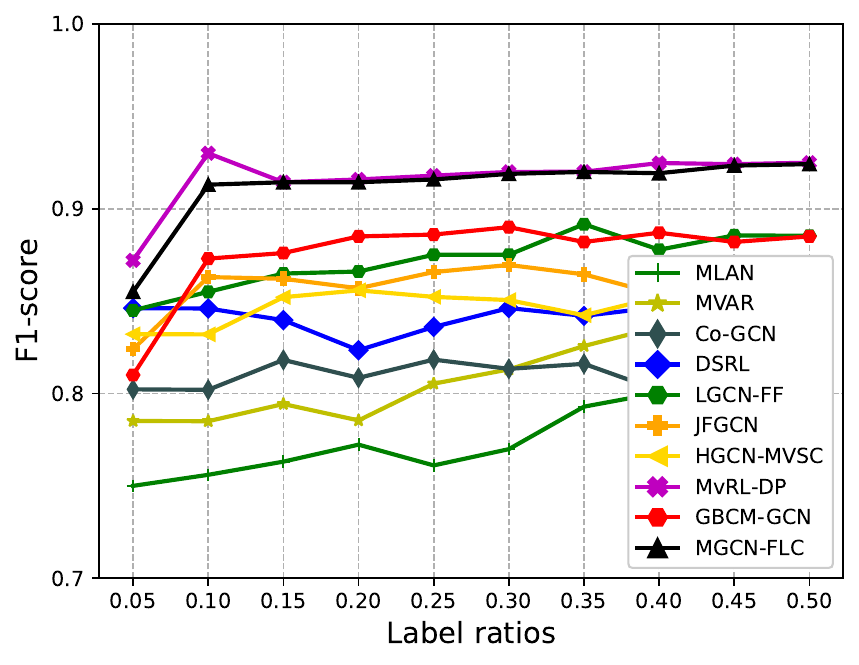}%
\label{fig_11_case}}
\vspace{-0.6cm}
\subfloat[\footnotesize MNIST]{\includegraphics[width=1.55in, trim=0cm 0cm 0.2cm 0.2cm clip]{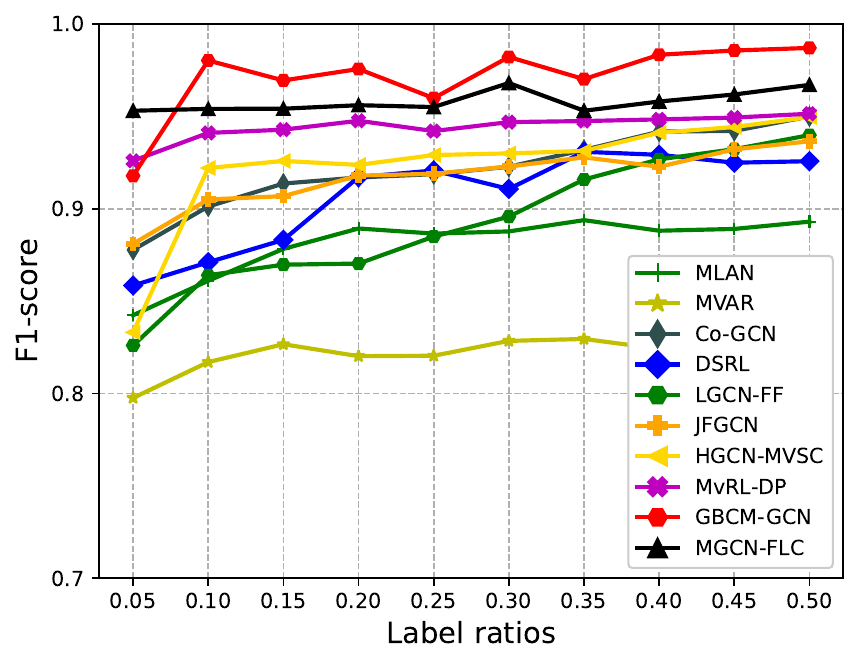}%
\label{fig_12_case}}
\vspace{1.2cm}
\subfloat[\footnotesize NUS-WIDE]{\includegraphics[width=1.55in, trim=0cm 0cm 0.2cm 0.2cm, clip]{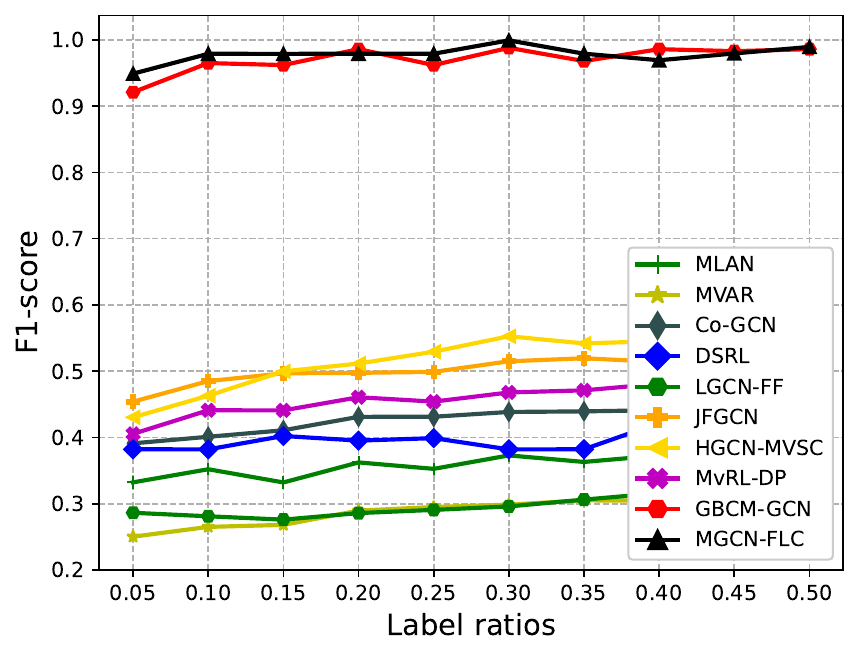}%
\label{fig_13_case}}
\vspace{-0.6cm}
\subfloat[\footnotesize WebKB]{\includegraphics[width=1.55in, trim=0cm 0cm 0.2cm 0.2cm, clip]{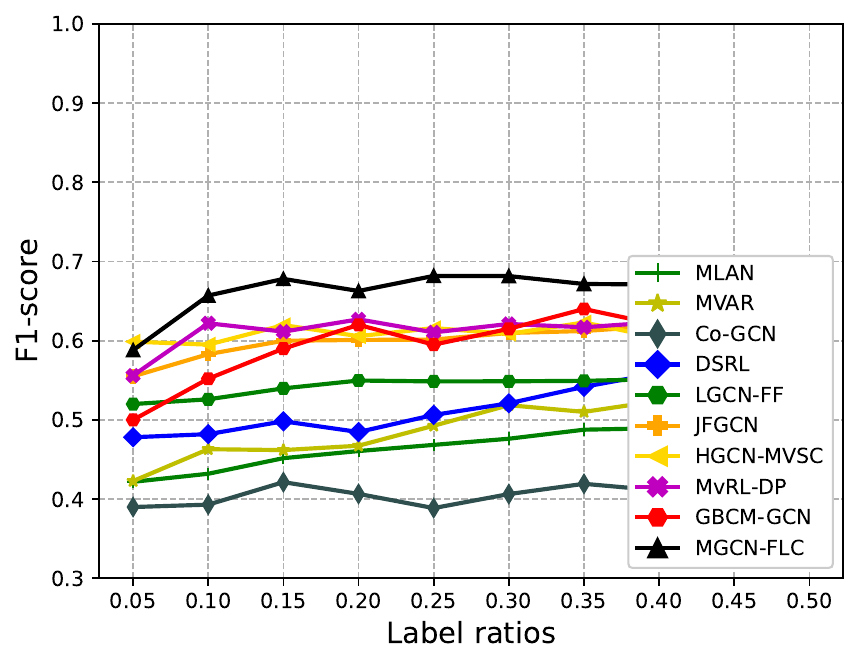}%
\label{fig_14_case}}
\vspace{-0.6cm}
\subfloat[\footnotesize BBCsports]{\includegraphics[width=1.55in, trim=0cm 0cm 0.2cm 0.2cm, clip]{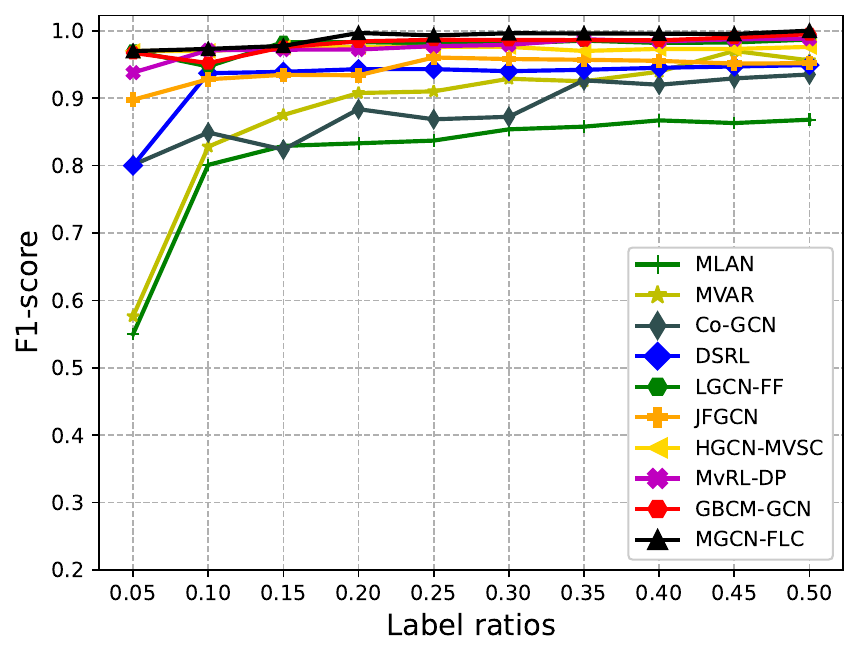}%
\label{fig_15_case}}
\vspace{1.1cm}
\subfloat[\footnotesize NGs]{\includegraphics[width=1.55in, trim=0cm 0cm 0.2cm 0.2cm, clip]{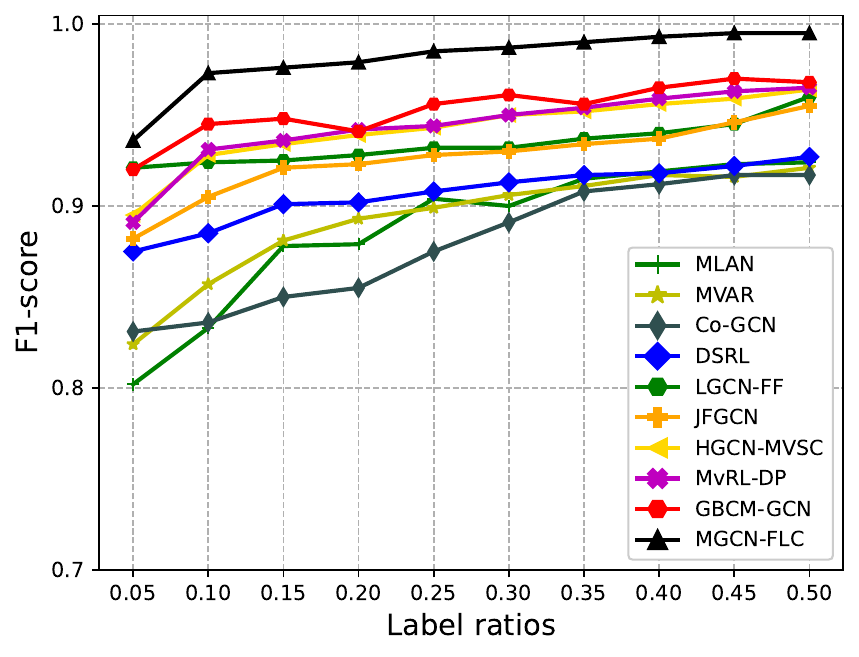}%
\label{fig_16_case}}
\vspace{-0.6cm}
\subfloat[\footnotesize ProteinFold ]{\includegraphics[width=1.55in, trim=0cm 0cm 0.2cm 0.2cm, clip]{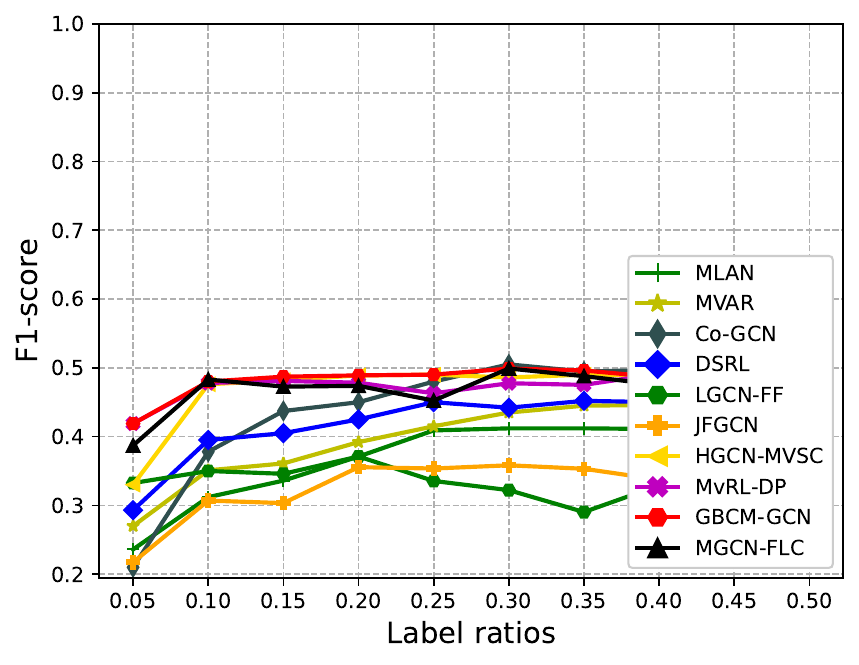}%
\label{fig_17_case}}
\vspace{-0.6cm}
\subfloat[\footnotesize Reuters]{\includegraphics[width=1.55in, trim=0cm 0cm 0.2cm 0.2cm, clip]{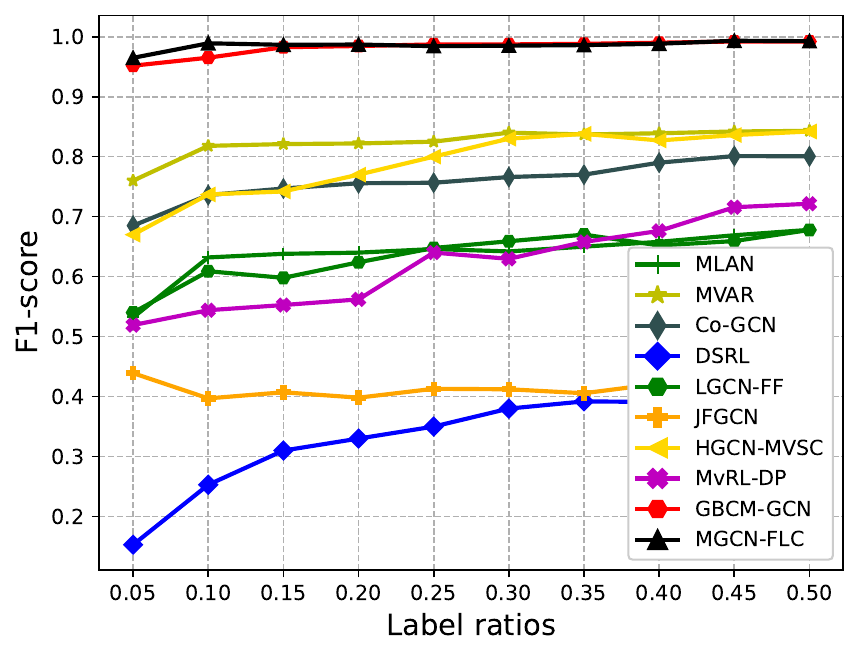}%
\label{fig_18_case}}
\vspace{1.1cm}
\caption{\small F1-score of all methods as the ratio of labeled data ranges in \{0.05, 0.10, \dots , 0.5\} on the nine datasets.}
\label{fig_sim2}
\end{figure*}

By leveraging feature interactions, MGCN-FLC captures inter-feature consistency and generates feature representations that incorporate such consistency. Compared with MvRL-DP \cite{DBLP:journals/kbs/WangLWGW25} and the GBCM-GCN \cite{DBLP:journals/eswa/WangYZXYX26}, which both utilize topology construction and cross-view interactive fusion, MGCN-FLC achieves further ACC improvements on most datasets. Furthermore, by explicitly and fully leveraging inter-view consistency, MGCN-FLC consistently outperforms that focus solely on topology construction, such as Co-GCN \cite{DBLP:conf/aaai/LiLW20a}, DSRL \cite{DBLP:journals/pami/WangCDL22}, LGCN-FF \cite{DBLP:journals/inffus/ChenFYGPW23}, JFGCN \cite{DBLP:journals/nn/ChenWCDW23}, and HGCN-MVSC \cite{DBLP:journals/nn/WangHWLCZ24}, demonstrating significant gains in ACC across all datasets.

\subsection{Classification performance with varying label ratios}
Figs. \ref{fig_sim1} and \ref{fig_sim2} show the ACC and F1-score of each algorithm under varying label ratios. Here, the label ratio refers to the proportion of labeled training samples relative to the total number of samples, ranging 5\% to 50\% in increments of 5\%. As shown in Fig. \ref{fig_sim1} and Fig. \ref{fig_sim2}, across all datasets, the ACC and F1-score of the various algorithms generally increase as the label ratio rises. MGCN-FLC performs stably and demonstrates a significant advantage over other methods, particularly at low label ratios.

\begin{figure*}[ht]
\centering
\subfloat[\footnotesize ACC]{\includegraphics[width=2.15in, trim=0cm 0cm 0.2cm 0.2cm, clip]{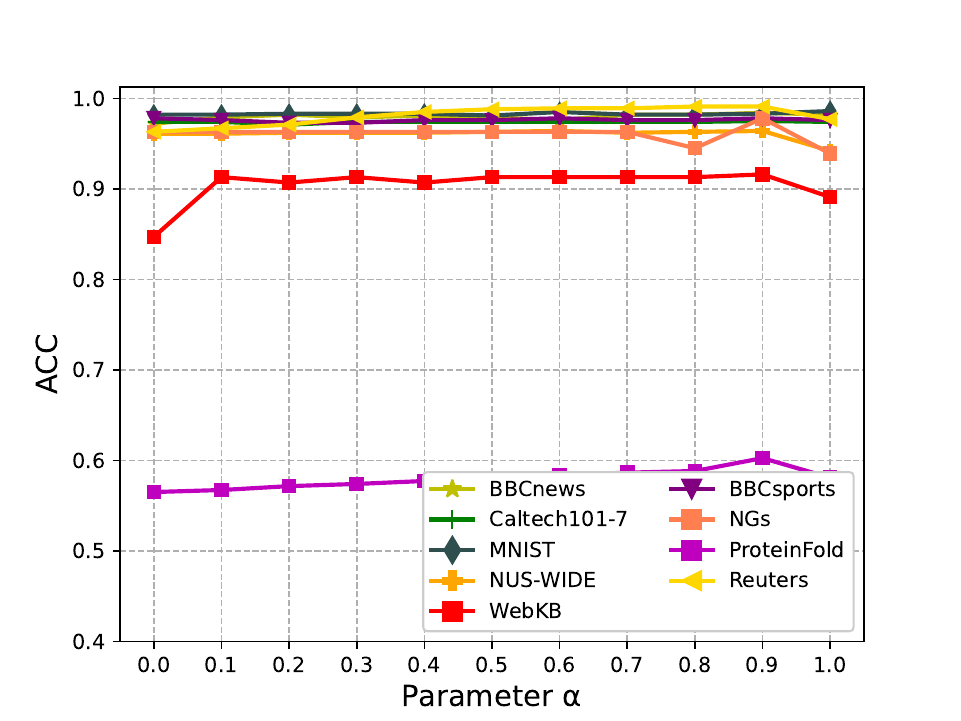}%
\label{α}}
\hfil
\subfloat [\footnotesize F1-score]{\includegraphics[width=2.15in, trim=0cm 0cm 0.2cm 0.2cm, clip]{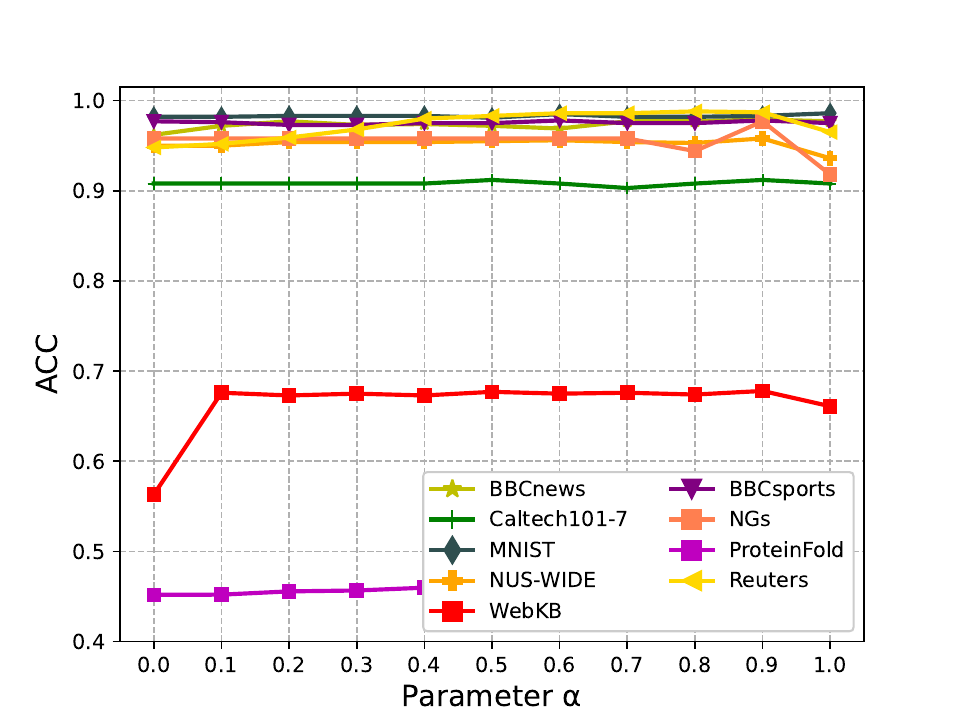}%
\label{α_f1}}
\caption{\small The ACC and F1-score of MGCN-FLC w.r.t. hyperparameter \(\alpha\) on datasets.}
\label{fig_α}
\end{figure*}

\begin{figure*}[ht]
\centering
\subfloat[\footnotesize ACC]{\includegraphics[width=2.15in, trim=0cm 0cm 0.2cm 0.2cm, clip]{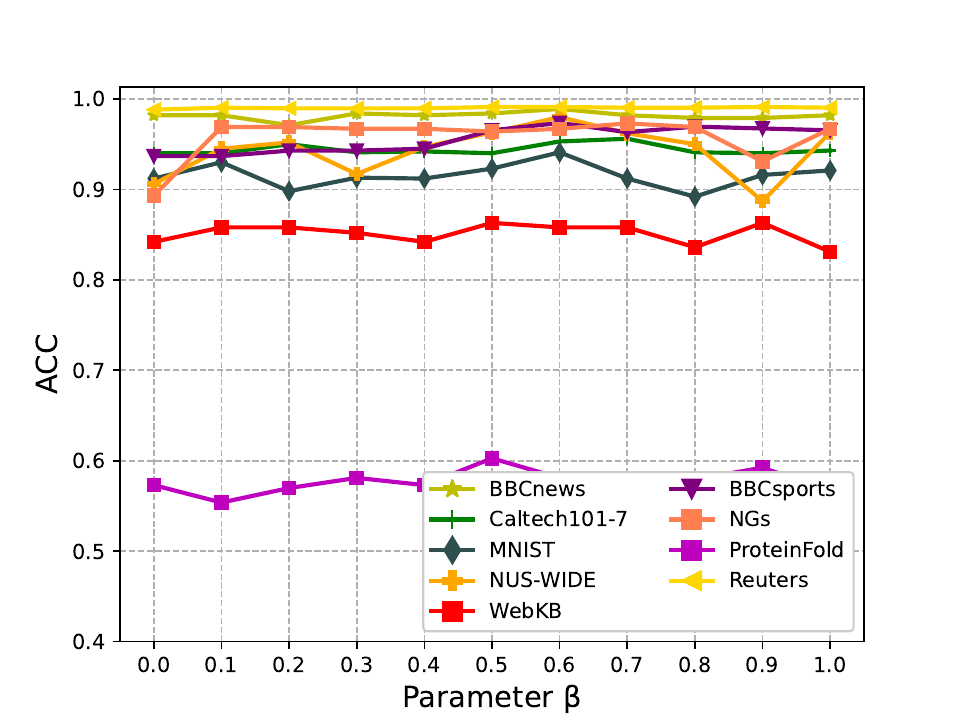}%
\label{fig20}}
\hfil
\subfloat [\footnotesize F1-score]{\includegraphics[width=2.15in, trim=0cm 0cm 0.2cm 0.2cm, clip]{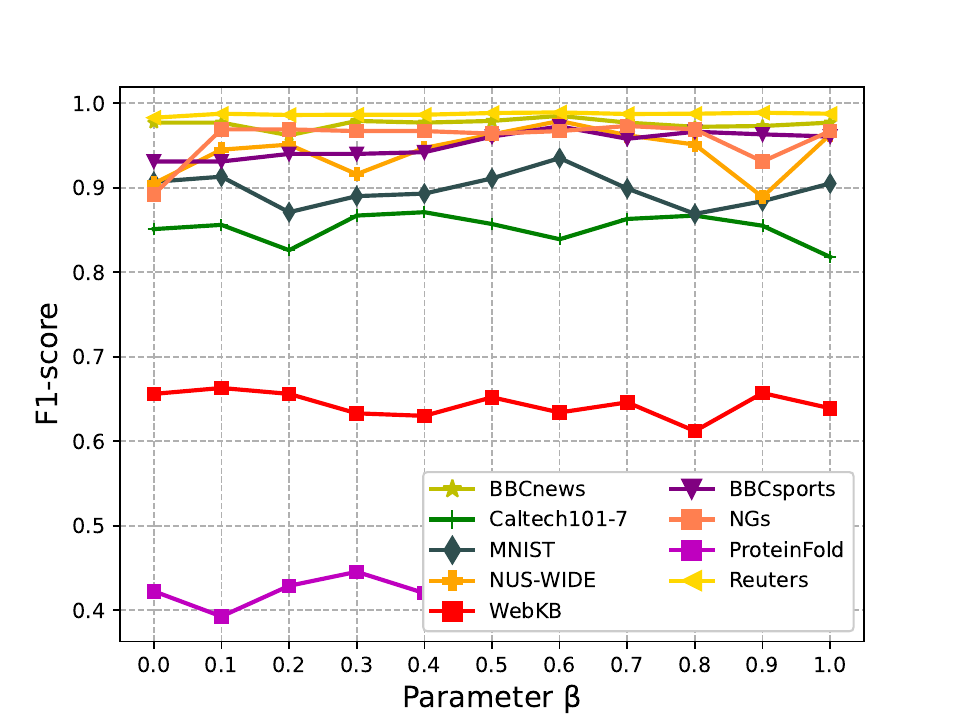}%
\label{fig21}}
\caption{\small The ACC and F1-score of MGCN-FLC w.r.t. hyperparameter \(\beta\) on datasets.}
\label{fig_sim4}
\end{figure*}

\subsection{Parameter necessity analysis}
The learnable hyperparameter weighting \(\alpha\) is used to control the fusion of intra-GB and inter-GB connections for constructing view-specific topology. The performance of MGCN-FLC under different values of \(\alpha\) is shown in Fig. \ref{fig_α}. It can be observed that the optimal ACC is achieved when \(\alpha\) is neither 0 nor 1, whereas relatively lower performance occurs when \(\alpha\) is set to 0 or 1. This observation indicates that the adaptive fusion of intra-GB and inter-GB connections is necessary. 

The hyperparameter \(\beta\) in the feature enhancement module is analyzed. Fig. \ref{fig_sim4} shows the performance of MGCN-FLC under different values of \(\beta\), with a range of \([0, 1]\) and a step size of 0.1. when \(\beta\) = 1, the feature enhancement module relies solely on max pooling, focusing exclusively on the salient information from the original feature matrix. Conversely, when \(\beta\) = 1, only average pooling is applied, emphasizing global information from the similarity feature matrix. As shown in Fig. \ref{fig_sim4}, the ACC reaches its best performance on the vast majority of datasets when \(\beta\) = 0.6, suggesting that the mixed pooling strategy effectively balances the information between the original feature matrix and similarity feature matrix. 

\begin{figure*}[ht]
\centering
\makebox[\textwidth][c]{%
{\color{red}\textbf{——}} Loss \quad 
{\color{blue}\textbf{——}} ACC \quad 
{\color{green}\textbf{——}} F1-score
}

\subfloat[\footnotesize BBCnews]{\includegraphics[width=1.5in, trim=0cm 0cm 0.2cm 0.2cm, clip]{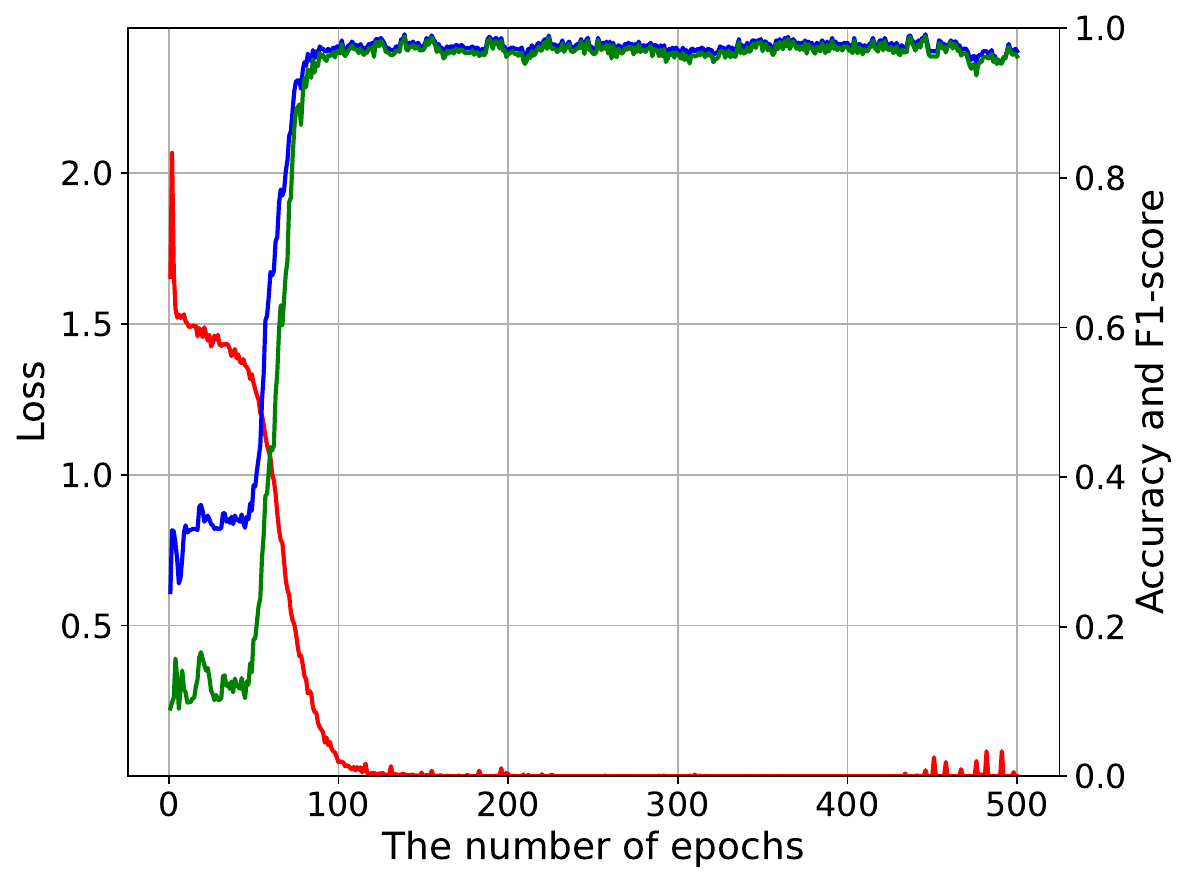}\label{BBCnews_loss}}%
\vspace{-0.6cm}
\subfloat[\footnotesize Caltech101-7]{\includegraphics[width=1.5in, trim=0cm 0cm 0.2cm 0.2cm, clip]{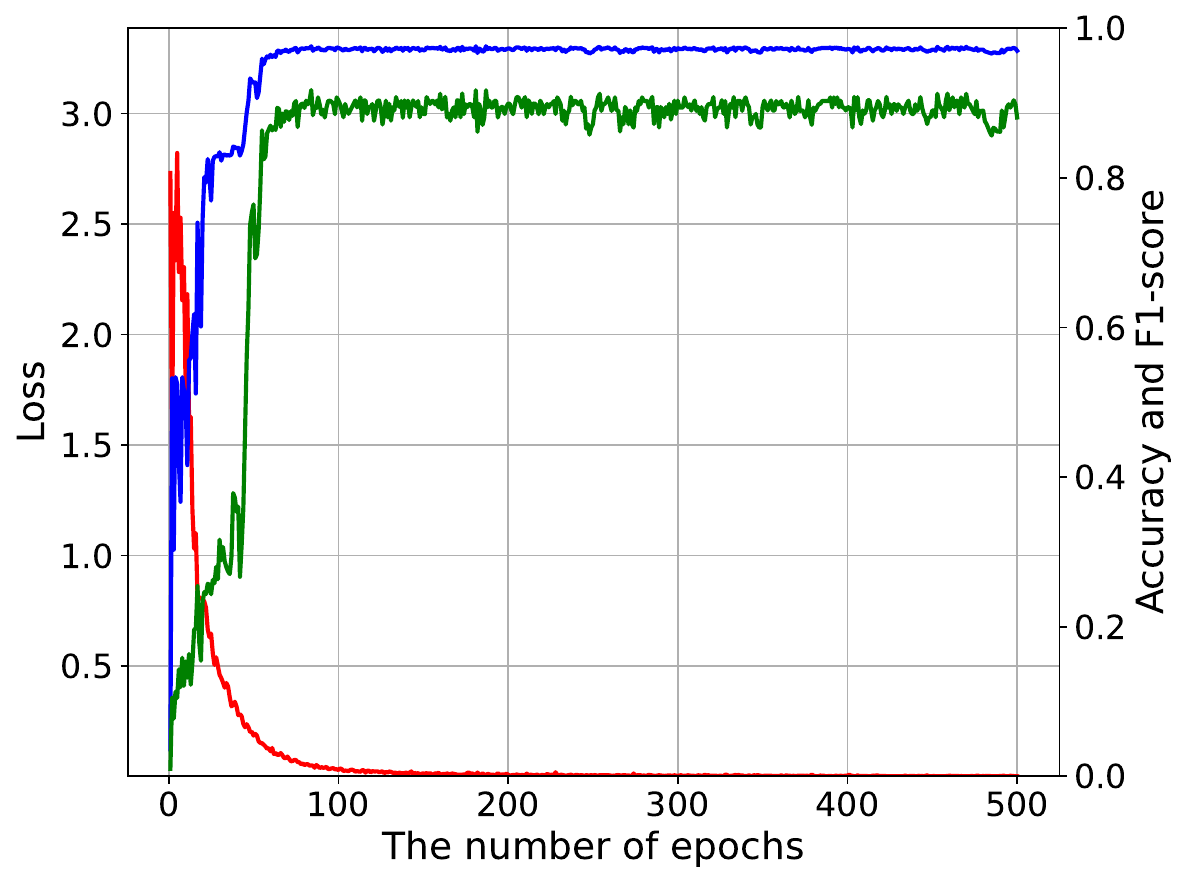}\label{Caltech101-7_loss}}
\vspace{1.2cm}
\subfloat[\footnotesize MNIST]{\includegraphics[width=1.5in, trim=0cm 0cm 0.2cm 0.2cm, clip]{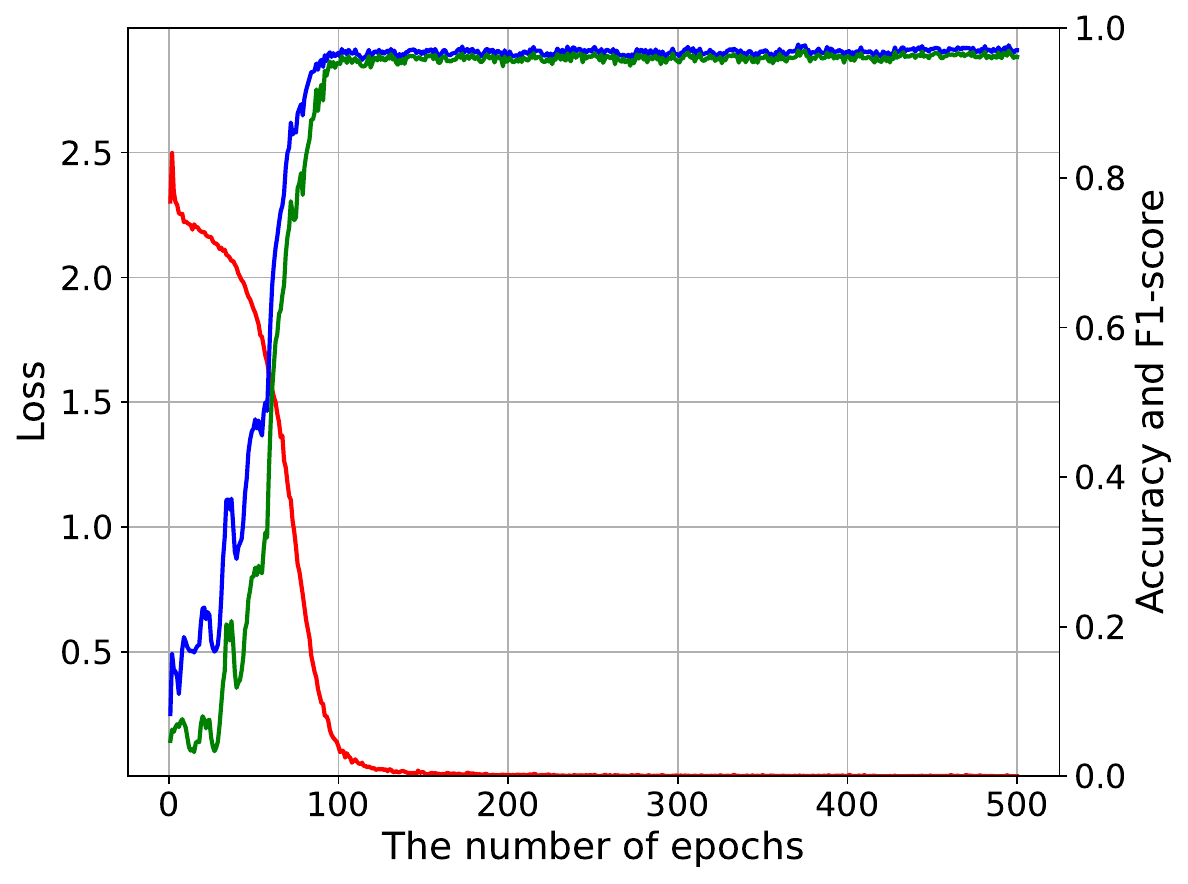}\label{MNIST_loss}}%
\vspace{-0.6cm}
\subfloat[\footnotesize NUS-WIDE]{\includegraphics[width=1.5in, trim=0cm 0cm 0.2cm 0.2cm, clip]{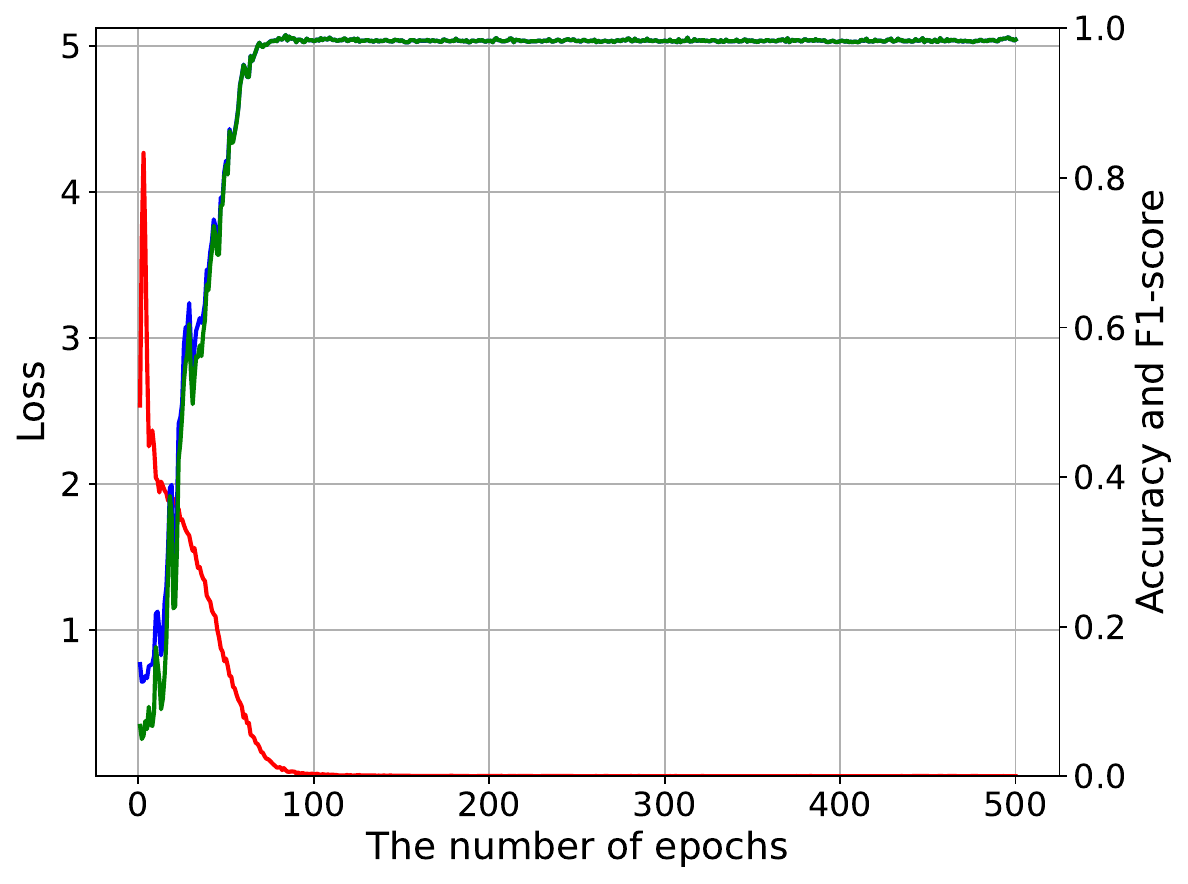}\label{NUS_WIDE_loss}}
\vspace{-0.6cm}
\subfloat[\footnotesize WebKB]{\includegraphics[width=1.5in, trim=0cm 0cm 0.2cm 0.2cm, clip]{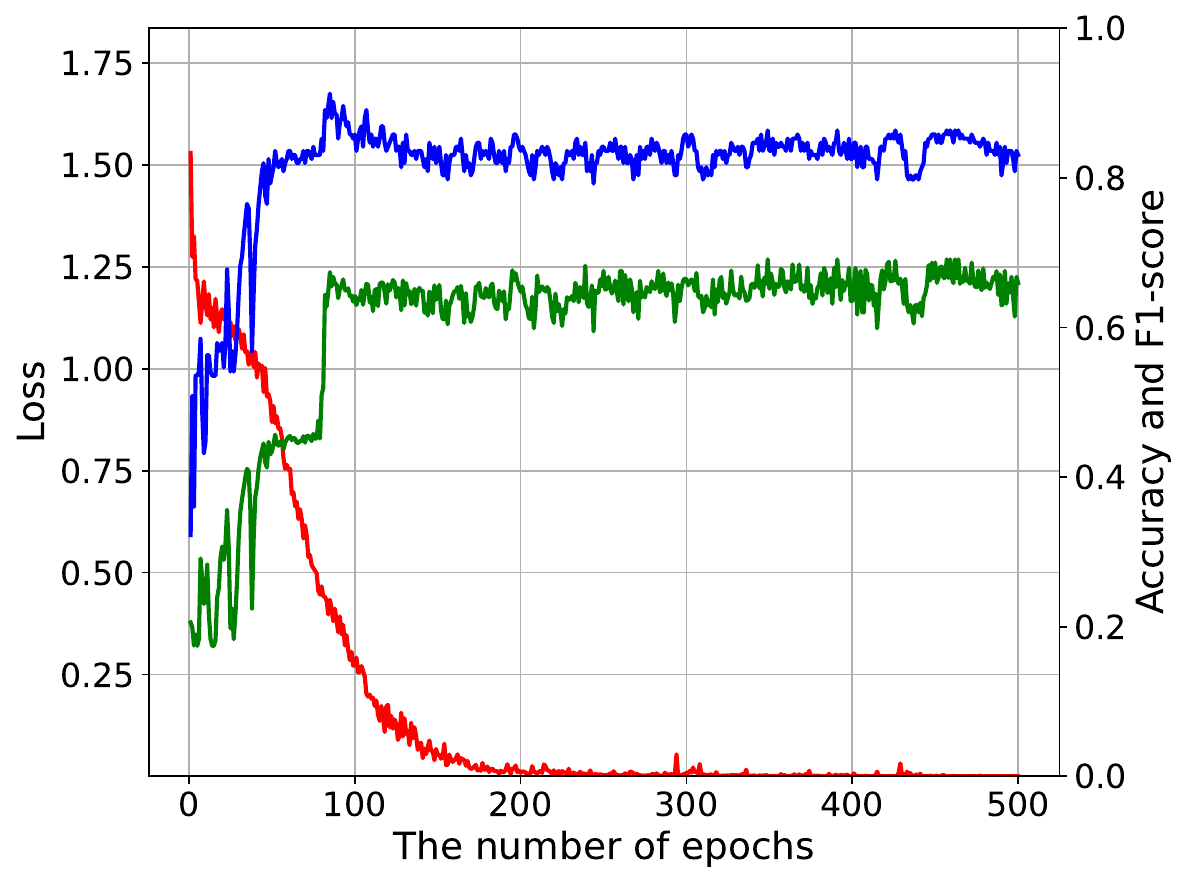}\label{WebKB_loss}}
\vspace{-0.6cm}
\subfloat[\footnotesize BBCsports]{\includegraphics[width=1.5in, trim=0cm 0cm 0.2cm 0.2cm, clip]{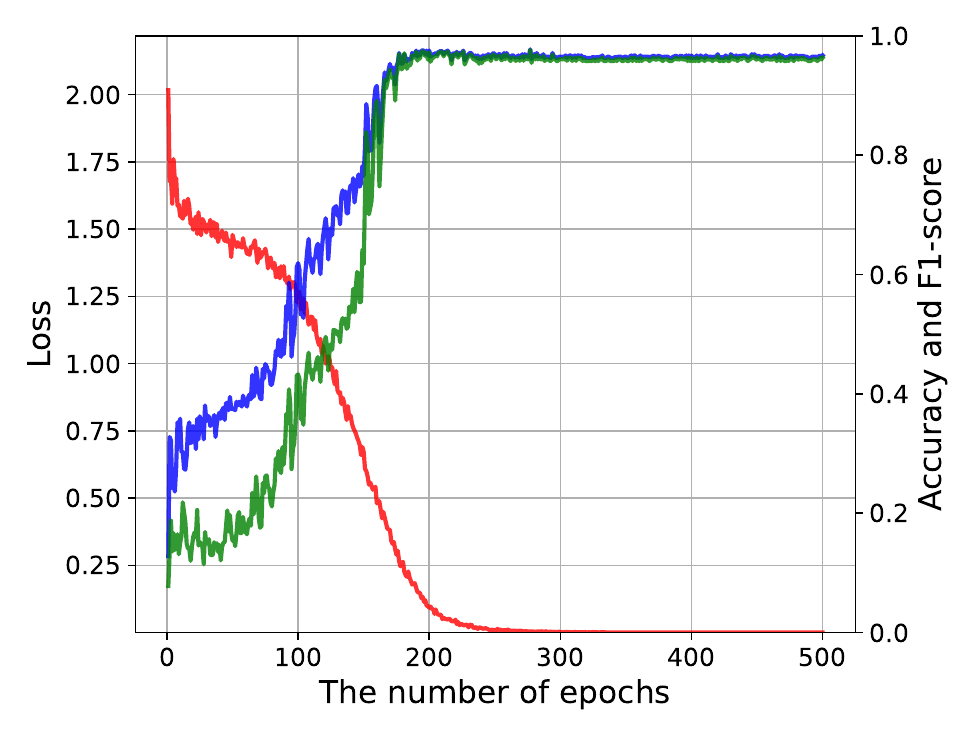}\label{BBCsports_loss}}
\vspace{1.1cm}
\subfloat[\footnotesize NGs]{\includegraphics[width=1.5in, trim=0cm 0cm 0.2cm 0.2cm, clip]{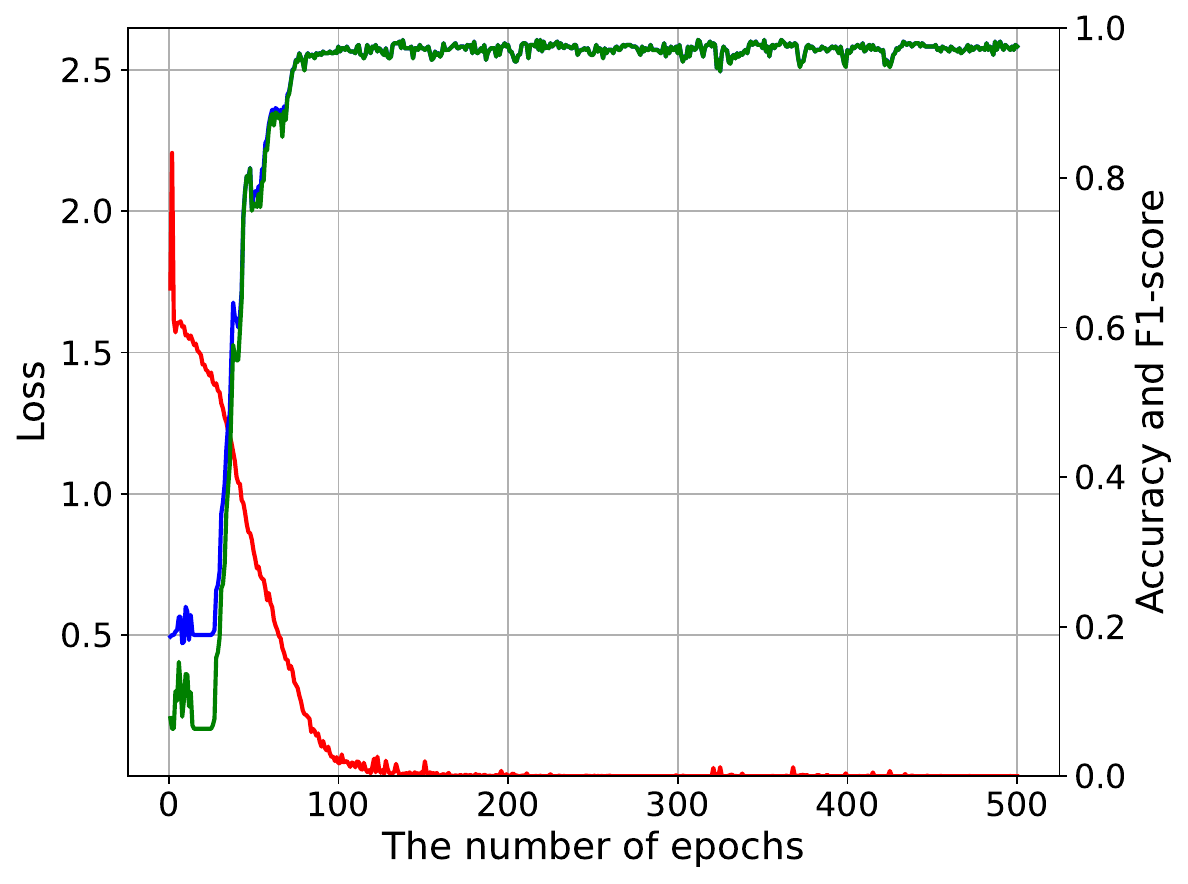}\label{NGs_loss}}
\vspace{-0.6cm}
\subfloat[\footnotesize ProteinFold]{\includegraphics[width=1.5in, trim=0cm 0cm 0.2cm 0.2cm, clip]{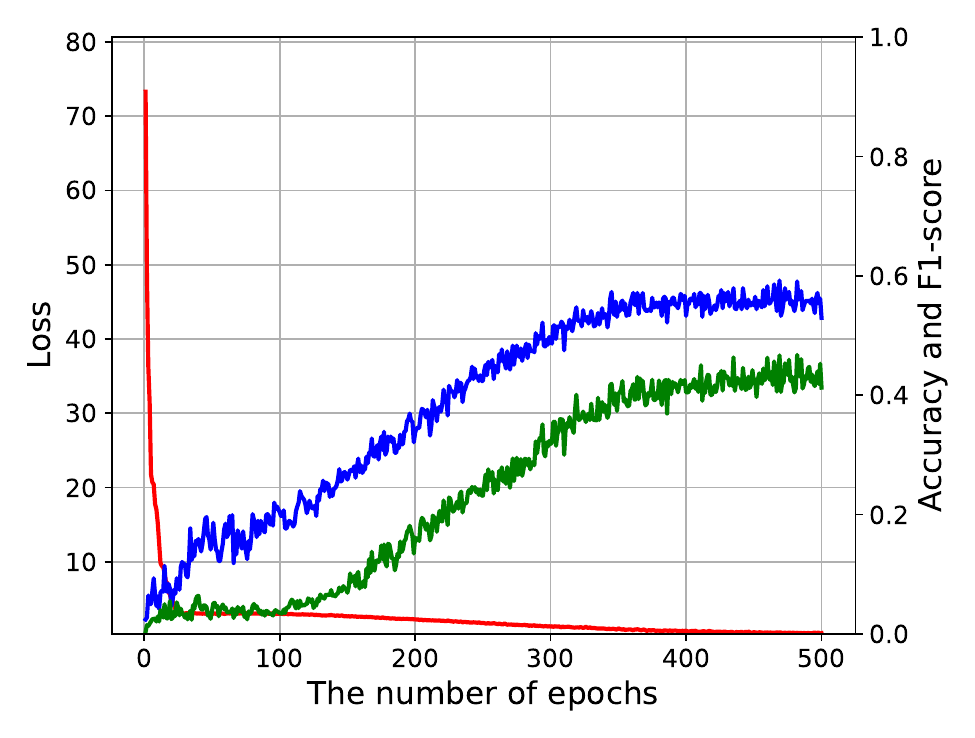}\label{ProteinFold_loss}}
\vspace{-0.6cm}
\subfloat[\footnotesize Reuters]{\includegraphics[width=1.5in, trim=0cm 0cm 0.2cm 0.2cm, clip]{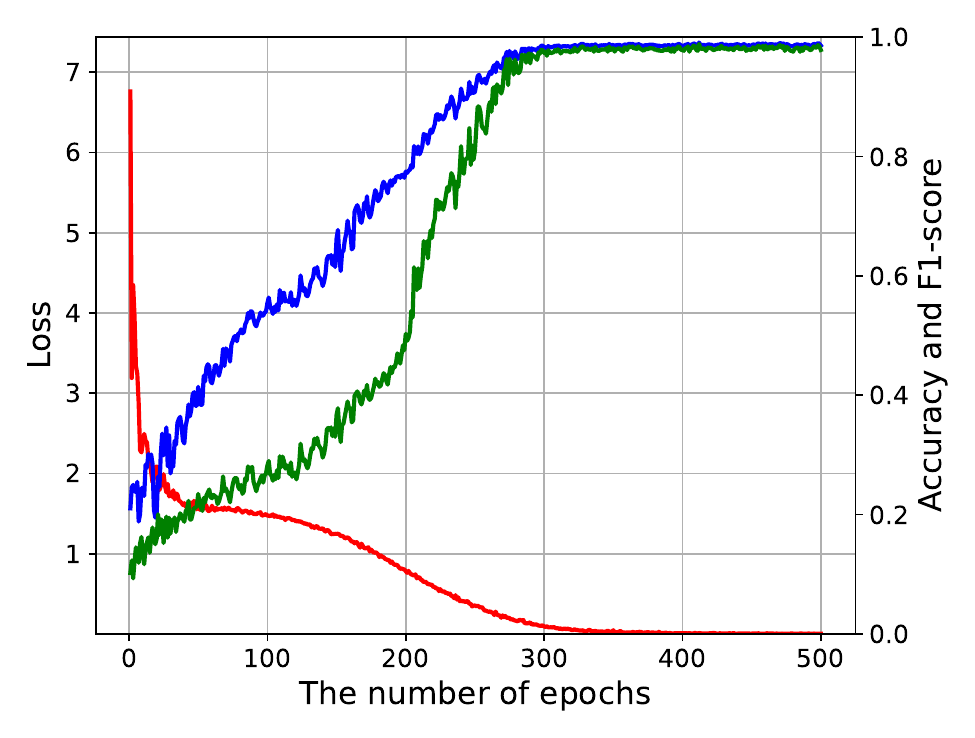}\label{Reuters_loss}}
\vspace{1.1cm}
\caption{\small The curves of MGCN-FLC’s loss values, ACC and F1-score on the selected datasets.}
\label{loss}
\end{figure*}

\subsection{Convergence analysis}
Fig. \ref{loss} illustrates the convergence process of MGCN-FLC across all datasets. The loss value decreases while the ACC\ textcolor{red}{and} F1-score increase until stabilization. The convergence process can be divided into two distinct phases: (1) the rapid convergence phase within the first 300 epochs, indicating effective learning, and (2) the stable convergence phase, demonstrating that the model parameters have been effectively optimized.

\begin{figure*}[ht]
\centering
\subfloat[\footnotesize JFGCN]{\includegraphics[width=1.5in, trim=0cm 0cm 0.2cm 0.2cm, clip]{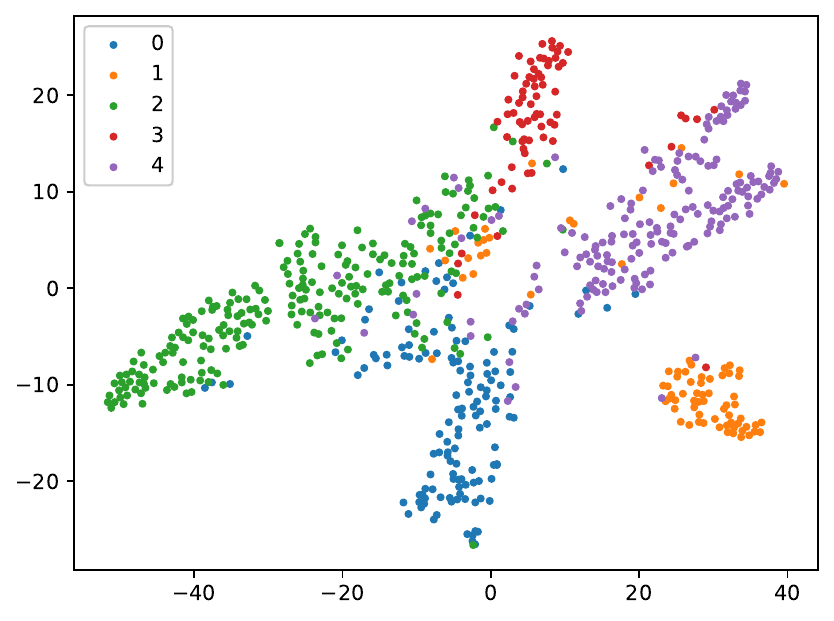}%
\label{fig_19_case}}
\hfil
\subfloat[\footnotesize LGCN-FF]{\includegraphics[width=1.5in, trim=0cm 0cm 0.2cm 0.2cm, clip]{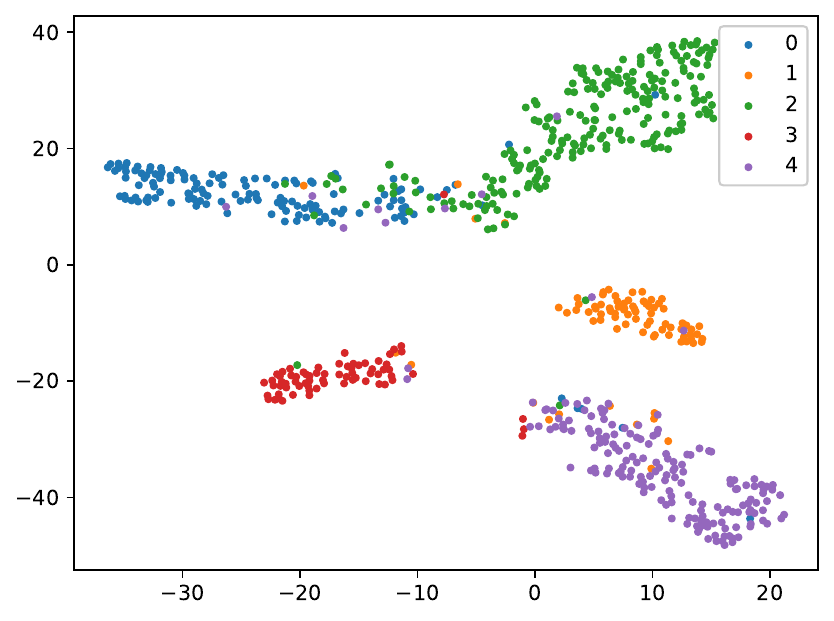}%
\label{fig_20_case}}
\hfil
\subfloat[\footnotesize MGCN-FLC]{\includegraphics[width=1.5in, trim=0cm 0cm 0.2cm 0.2cm, clip]{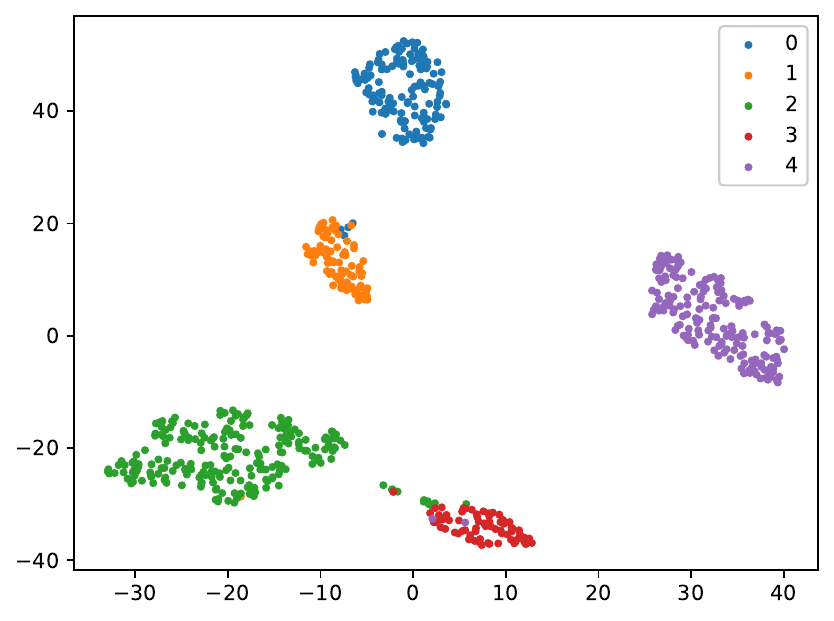}%
\label{fig_21_case}}
\hfil
\subfloat[\footnotesize JFGCN]{\includegraphics[width=1.5in, trim=0cm 0cm 0.2cm 0.2cm, clip]{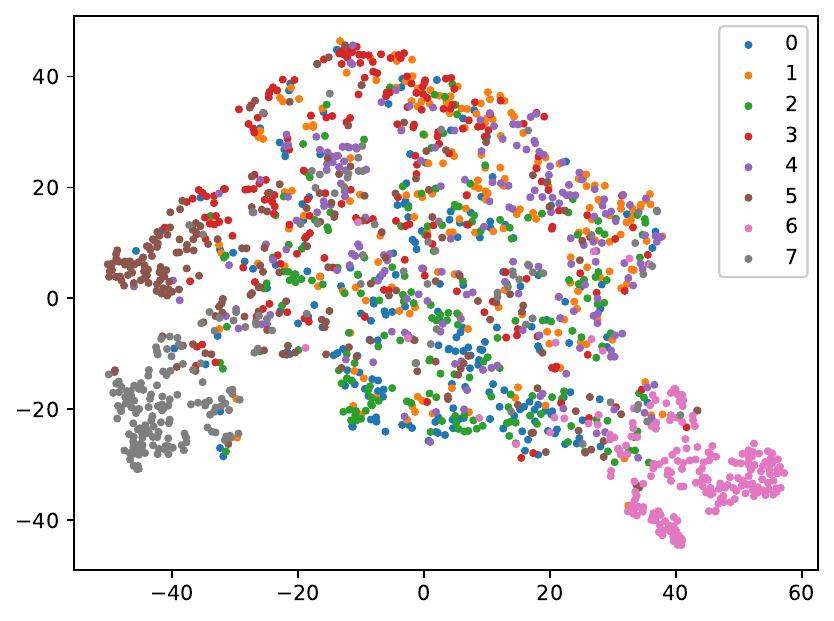}%
\label{fig22}}
\hfil
\subfloat[\footnotesize LGCN-FF]{\includegraphics[width=1.5in, trim=0cm 0cm 0.2cm 0.2cm, clip]{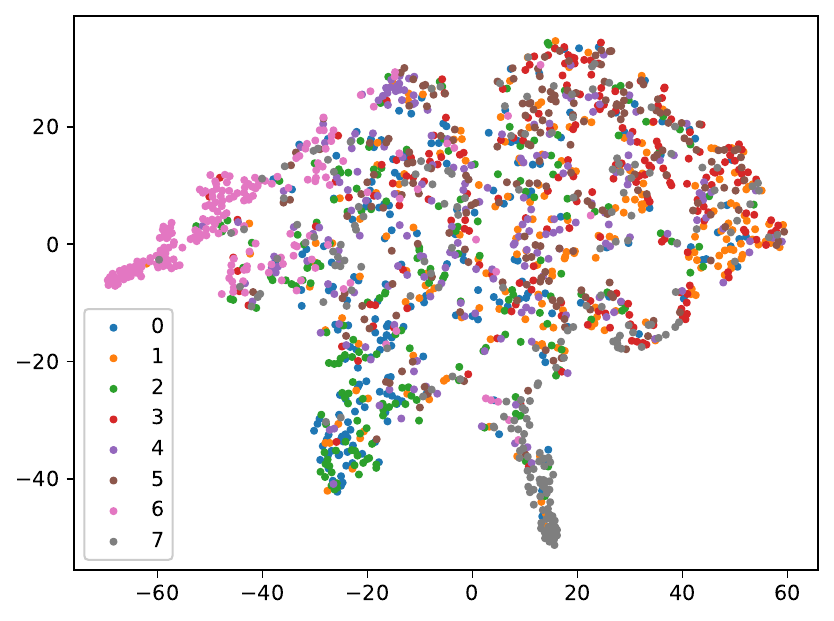}%
\label{fig23}}
\hfil
\subfloat[\footnotesize MGCN-FLC]{\includegraphics[width=1.5in, trim=0cm 0cm 0.2cm 0.2cm, clip]{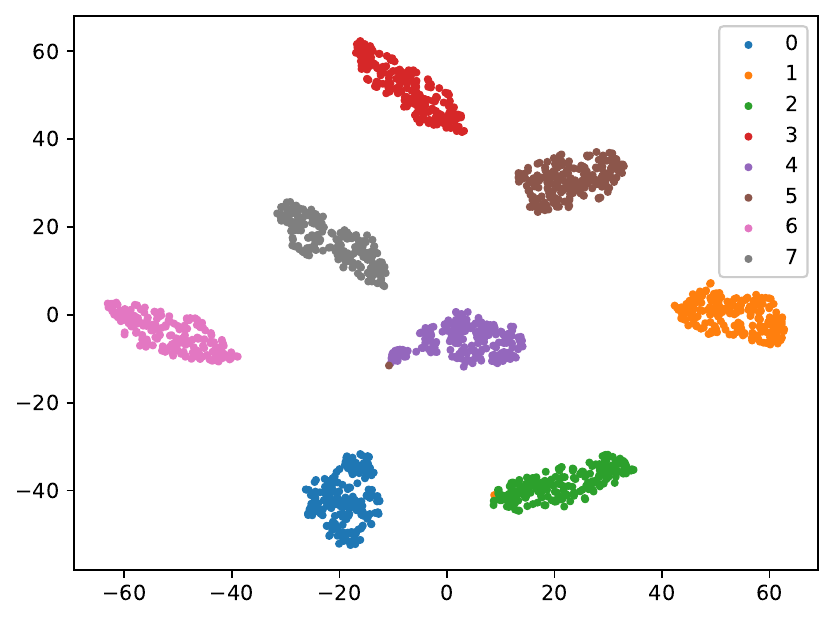}%
\label{fig24}}
\caption{\small t-SNE visualizations of the features from JFGCN, LGCN-FF, and MGCN-FLC on the (a-c) BBCnews and (d-f) NUS-WIDE datasets, with colors denoting different classes.}
\label{fig5}
\end{figure*}

\subsection{Visualization analysis}
Fig. \ref{fig5} visualizes the node embeddings projected onto a two-dimensional plane using the tSNE method, showcasing the effectiveness of node classification. We present the visualization results of node embeddings obtained by JFGCN, LGCN-FF, and MGCN-FLC on the BBCnews and NUS-WIDE datasets, respectively. Our MGCN-FLC algorithm makes the nodes of the same category to cluster together, while nodes of different categories are effectively separated. Although some outlier nodes remain, the number of them is decreased compared to other methods. The visualization results can also demonstrate that MGCN-FLC outperforms other algorithms in semi-supervised node classification tasks.

\subsection{Ablation experiment}

\subsubsection{Contribution of main modules}
Ablation experiments are conducted to assess the contribution of each of the three modules in MGCN-FLC model. For clarity, the topology construction module is denoted as \(tc\), the feature enhancement module as \(fe\), and the interactive fusion module as (\(if\)). The specific ablation settings are as follows:
\begin{itemize}
 \item{\bf  \(tc\)}: {The \textit{k}NN  algorithm replaces the unsupervised GB algorithm to evaluate the contribution of the GB-based topology construction module to the classification performance.}  
 \item{\bf  \(fe\)}: {The original feature is used to replace the enhanced feature to assess the contribution of the feature enhancement module.}
 \item{\bf  \(if\)}: {Traditional multi-view fusion method (i.e. weighted sum) replaces the interactive fusion method to evaluate its contribution to interactive fusion module to the classification performance.}
 \end{itemize}

\begin{table}[ht]
\caption{Ablation study results of the proposed MGCN-FLC on all datasets.}\label{tab:table2}
    \centering
{\fontsize{6}{7}\selectfont
    \begin{tabular}{l|ccc|c}
\noalign{\hrule height 1pt}
Datasets/Methods & $\textbf{MGCN-FLC}_{\text{tc}}$ & $\textbf{MGCN-FLC}_{\text{fe}}$ & $\textbf{MGCN-FLC}_{\text{if}}$ &  $\textbf{MGCN-FLC}$\\
 \hline
\begin{tabular}{p{1.5cm}p{0.5cm}}BBCnews& \begin{tabular}{l}ACC \\ F1\end{tabular} \\\end{tabular}  &  \begin{tabular}{c}91.1 \\90.2 \end{tabular} & \begin{tabular}{c}95.3 \\94.3\end{tabular} & \begin{tabular}{c}{\bf 98.4} \\{\bf97.9}\end{tabular} &  \begin{tabular}{c}{\bf 98.4} \\{\bf97.9}\end{tabular}\\
\hline
\begin{tabular}{p{1.5cm}p{0.5cm}}Caltech101-7& \begin{tabular}{l}ACC \\ F1\end{tabular} \\\end{tabular}  & \begin{tabular}{c}83.8\\73.1\end{tabular} & \begin{tabular}{c}92.0 \\83.3\end{tabular} &  \begin{tabular}{c}97.4 \\90.7\end{tabular} &  \begin{tabular}{c}{\bf97.5} \\{\bf91.3}\end{tabular}\\
\hline
\begin{tabular}{p{1.5cm}p{0.5cm}}MNIST& \begin{tabular}{l}ACC \\ F1\end{tabular} \\\end{tabular}  &  \begin{tabular}{c}90.6 \\88.6\end{tabular} & \begin{tabular}{c}95.0 \\94.3\end{tabular} & \begin{tabular}{c}82.9 \\76.9\end{tabular} &  \begin{tabular}{c}{\bf96.4 }\\{\bf95.8}\end{tabular}\\
 \hline
\begin{tabular}{p{1.5cm}p{0.5cm}}NUS-WIDE& \begin{tabular}{l}ACC \\ F1\end{tabular} \\\end{tabular}    & \begin{tabular}{c}40.9 \\38.9\end{tabular} & \begin{tabular}{c}98.1 \\98.1\end{tabular} & \begin{tabular}{c}98.1 \\98.1\end{tabular} &  \begin{tabular}{c}{\bf99.0} \\{\bf99.0}\end{tabular}\\
 \hline
\begin{tabular}{p{1.5cm}p{0.5cm}}WebKB& \begin{tabular}{l}ACC \\ F1\end{tabular} \\\end{tabular}  &  \begin{tabular}{c}82.0 \\44.6\end{tabular} & \begin{tabular}{c}86.9 \\66.3\end{tabular} & \begin{tabular}{c}{\bf91.8 }\\{\bf67.4}\end{tabular} &  \begin{tabular}{c}{\bf91.8 }\\{\bf67.4}\end{tabular}\\
\hline
\begin{tabular}{p{1.5cm}p{0.5cm}}BBCsports& \begin{tabular}{l}ACC \\ F1\end{tabular} \\\end{tabular}  &  \begin{tabular}{c}88.6 \\89.6\end{tabular} & \begin{tabular}{c}93.5 \\93.6\end{tabular} & \begin{tabular}{c}96.3 \\95.8\end{tabular} &  \begin{tabular}{c}{\bf97.3 }\\{\bf97.2}\end{tabular}\\
\hline
\begin{tabular}{p{1.5cm}p{0.5cm}}NGs& \begin{tabular}{l}ACC \\ F1\end{tabular} \\\end{tabular}  &  \begin{tabular}{c}94.2 \\94.3\end{tabular} & \begin{tabular}{c}93.8 \\93.8\end{tabular} & \begin{tabular}{c}{\bf97.8 }\\{\bf97.8}\end{tabular} &  \begin{tabular}{c}{\bf97.8 }\\{\bf97.8}\end{tabular}\\
\hline
\begin{tabular}{p{1.5cm}p{0.5cm}}ProteinFold
& \begin{tabular}{l}ACC \\ F1\end{tabular} \\\end{tabular}  &  \begin{tabular}{c}31.0 \\15.4\end{tabular} & \begin{tabular}{c}39.6 \\20.0\end{tabular} & \begin{tabular}{c}29.4 \\11.9\end{tabular} &  \begin{tabular}{c}{\bf60.0 }\\{\bf48.3}\end{tabular}\\
\hline
\begin{tabular}{p{1.5cm}p{0.5cm}}Reuters
& \begin{tabular}{l}ACC \\ F1\end{tabular} \\\end{tabular}  &  \begin{tabular}{c}72.7 \\68.8\end{tabular} & \begin{tabular}{c}87.0 \\63.2\end{tabular} & \begin{tabular}{c}98.1\\96.6\end{tabular} &  \begin{tabular}{c}{\bf99.1 }\\{\bf98.9}\end{tabular}\\
\noalign{\hrule height 1pt}
    \end{tabular}
\label{table 4}}
\end{table}

\begin{figure*}[ht]
\centering
\includegraphics[width=1.0\textwidth]{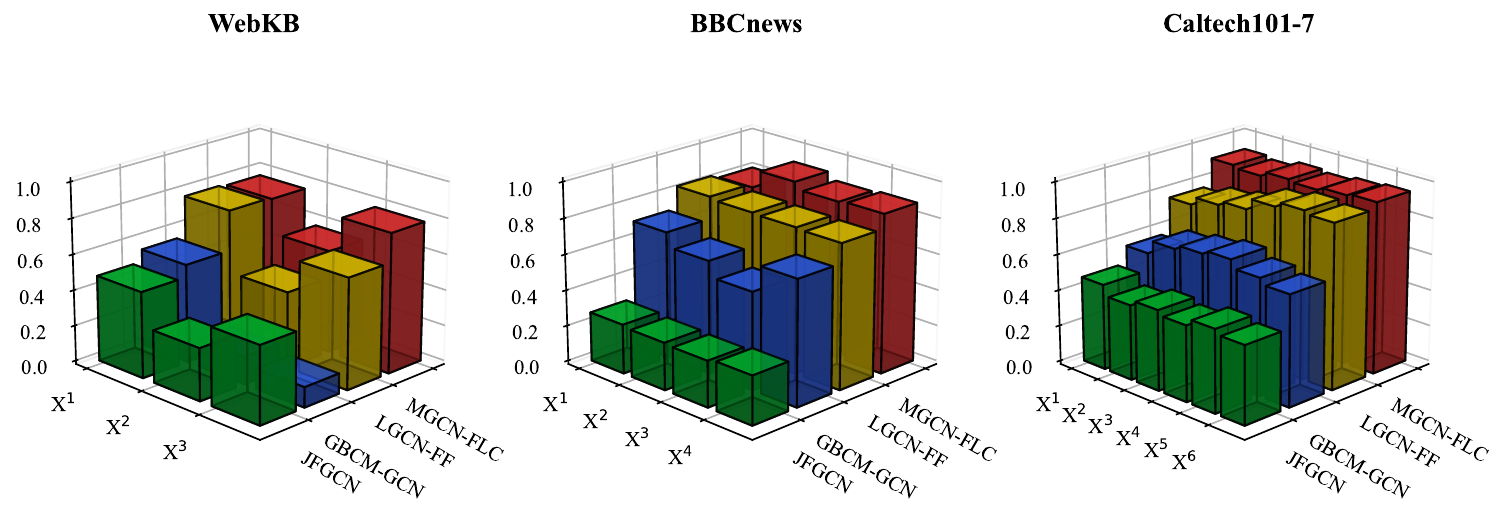}
\caption{The homophily ratios of the topologies generated by the four algorithms—JFGCN, GBCM-GCN, LGCN-FF, and MGCN-FLC—across the WebKB, BBCnews, and Caltech101-7 datasets.}
\label{fig18}
\end{figure*}
The results of the ablation experiments, including the classification performance (ACC and F1-scores) with each module ablated, are presented in Table \ref{table 4}. 

Ablating the \(tc\) module leads to varying degrees of classification performance degradation across all datasets for the MGCN-FLC method, with a particularly notable ACC drop of 58.1\% on the NUS-WIDE dataset. This ablation highlights the contribution of the GB-based topology construction module which not only avoids the inherent \textit{k}-value noise in \textit{k}NN but also achieves the high homophily ratio (as shown in Fig. \ref{fig18}), enabling MGCN-FLC to construct high-quality topology. 

Ablating the \(fe\) module also results in decreased classification performance across all datasets for the MGCN-FLC method, with significant ACC drop of 5.5\% on the Caltech101-7 dataset. This demonstrates the importance of the feature enhancement module, which leverages inter-feature consistency to enable MGCN-FLC to generate feature representations. 

Ablating the \(if\) module, the ablation of it leads to degraded classification performance of the MGCN-FLC method on some datasets, notably with ACC drop of 13.5\% on the MNIST dataset. This confirms the importance of the interactive fusion module in explicitly and fully exploiting inter-view consistency. It is worth noting, however, that on the BBCnews, WebKB, and NGs datasets, the ACC remains largely unchanged after ablating the \(if\) module. This occurs because the feature representations of different views in these datasets are highly similar in terms of classification ability, resulting in feature representations produced by the interactive fusion module that converge with those from a single view. For instance, on the WebKB dataset, the individual views have achieve accuracies of 91.2\%, 91.2\%, and 91.7\%, respectively. Regardless of whether the \(if\) module is used, MGCN-FLC achieves the ACC of 91.8\%. This indicates that the powerful classification capability of the feature representations from the individual views leaves limited room for the \(if\) module to demonstrate its advantages. But there's no need to worry, because the powerful classification capability of the feature representations from the individual views actually originates from the other two modules of the MGCN-FLC method, i.e. \(tc\) and \(fe\).

\subsubsection{The superiority of representative node selection}
\begin{table}[ht]
\caption{The classification results of MGCN-FLC based on representative node and its variants (center node and density peak node) on all datasets.}\label{tab:table3}
    \centering
{\fontsize{6}{7}\selectfont
    \begin{tabular}{l|c|c|c}
\noalign{\hrule height 1pt}
\multirow{2}{*}{\textbf{Datasets/Methods}} & \multicolumn{3}{c}{\textbf{MGCN-FLC}} \\ \cline{2-4}
& \textbf{center node} & \textbf{density peak node} & \textbf{representative node (our)} \\
 \hline
\begin{tabular}{p{1.5cm}p{0.5cm}}BBCnews& \begin{tabular}{l}ACC \\ F1\end{tabular} \\\end{tabular}  &  \begin{tabular}{c}{\bf98.4}\\{\bf97.9}\end{tabular} & \begin{tabular}{c}{\bf98.4} \\{\bf97.9}\end{tabular} &  \begin{tabular}{c}{\bf 98.4} \\{\bf97.9}\end{tabular}\\
\hline
\begin{tabular}{p{1.5cm}p{0.5cm}}Caltech101-7& \begin{tabular}{l}ACC \\ F1\end{tabular} \\\end{tabular}   & \begin{tabular}{c}{\bf97.5} \\{\bf91.3}\end{tabular} &  \begin{tabular}{c}{\bf97.5} \\{\bf91.3}\end{tabular} &  \begin{tabular}{c}{\bf97.5} \\{\bf91.3}\end{tabular}\\
\hline
\begin{tabular}{p{1.5cm}p{0.5cm}}MNIST& \begin{tabular}{l}ACC \\ F1\end{tabular} \\\end{tabular}  & \begin{tabular}{c}95.4 \\94.8\end{tabular} & \begin{tabular}{c}\underline{95.9}\\\underline{94.9}\end{tabular} &  \begin{tabular}{c}{\bf96.4 }\\{\bf95.8}\end{tabular}\\
 \hline
\begin{tabular}{p{1.5cm}p{0.5cm}}NUS-WIDE& \begin{tabular}{l}ACC \\ F1\end{tabular} \\\end{tabular}   & \begin{tabular}{c}{\bf99.0} \\{\bf99.0}\end{tabular} & \begin{tabular}{c}\underline{98.5} \\\underline{98.5}\end{tabular} &  \begin{tabular}{c}{\bf99.0} \\{\bf99.0}\end{tabular}\\
 \hline
\begin{tabular}{p{1.5cm}p{0.5cm}}WebKB& \begin{tabular}{l}ACC \\ F1\end{tabular} \\\end{tabular}  & \begin{tabular}{c}{\bf91.8} \\{\bf67.4}\end{tabular} & \begin{tabular}{c}{\bf91.8} \\{\bf67.4}\end{tabular} &  \begin{tabular}{c}{\bf91.8 }\\{\bf67.4}\end{tabular}\\
\hline
\begin{tabular}{p{1.5cm}p{0.5cm}}BBCsports& \begin{tabular}{l}ACC \\ F1\end{tabular} \\\end{tabular} & \begin{tabular}{c}{\bf97.3} \\{\bf97.2}\end{tabular} & \begin{tabular}{c}{\bf97.3} \\{\bf97.2}\end{tabular} &  \begin{tabular}{c}{\bf97.3 }\\{\bf97.2}\end{tabular}\\
\hline
\begin{tabular}{p{1.5cm}p{0.5cm}}NGs& \begin{tabular}{l}ACC \\ F1\end{tabular} \\\end{tabular} & \begin{tabular}{c}\underline{97.4}\\\underline{97.4}\end{tabular} & \begin{tabular}{c}{\bf97.8}\\{\bf97.8}\end{tabular} &  \begin{tabular}{c}{\bf97.8 }\\{\bf97.8}\end{tabular}\\
\hline
\begin{tabular}{p{1.5cm}p{0.5cm}}ProteinFold
& \begin{tabular}{l}ACC \\ F1\end{tabular} \\\end{tabular}  & \begin{tabular}{c}\underline{58.7} \\\underline{47.8}\end{tabular} & \begin{tabular}{c}58.3 \\46.0\end{tabular} &  \begin{tabular}{c}{\bf60.0 }\\{\bf48.3}\end{tabular}\\
\hline
\begin{tabular}{p{1.5cm}p{0.5cm}}Reuters
& \begin{tabular}{l}ACC \\ F1\end{tabular} \\\end{tabular}  & \begin{tabular}{c}{\bf99.1} \\{\bf98.9}\end{tabular} & \begin{tabular}{c}{\bf99.1} \\{\bf98.9}\end{tabular} &  \begin{tabular}{c}{\bf99.1 }\\{\bf98.9}\end{tabular}\\
\noalign{\hrule height 1pt}
    \end{tabular}
\label{table 5}}
\end{table}
In topology construction, we select representative nodes by “minimizing the sum of Euclidean distances to other nodes”. To validate the superiority of the selection approach, we replaced the original representative node with center node and density peak node, respectively. In detail, the center node is selected via the mean vector of nodes within the GB, and the density peak node is selected according to the highest density within the GB. Table \ref{table 5} presents the classification performance of the three methods based on different representative node selection strategies on all datasets. Compared with the variant based on center node, our MGCN-FLC improves ACC by 1\%, 0.4\%, and 1.3\% on the MNIST, NGs, and ProteinFold datasets, respectively. These improvements are attributed to MGCN-FLC's strategy of selecting actual nodes and focusing more on the node itself, while the virtual nature of the center node may introduce noise. Compared with the variant based on density peak node, MGCN-FLC improves ACC by 0.5\%, 0.5\%, and 1.7\% on the MNIST, NUS-WIDE, and ProteinFold datasets, respectively. This is because MGCN-FLC does not require parameter tuning, whereas adjustments to parameters may affect the selection of density peak nodes, potentially leading to connections between dissimilar nodes. The identical classification results observed on datasets such as BBCnews and Caltech101-7 can be attributed to the similar features of the selected nodes, which consequently do not affect the connections between similar GBs.

\subsubsection{Advangtage of interactive fusion module}
\begin{table}[ht]
\caption{The classification results of MGCN-FLC and its variants on all datasets. 'VA-based' denotes the MGCN-FLC variant based on view-level attention. 'VSF-based' denotes the MGCN-FLC variant based on view-shared underlying features. 'IF-based' denotes the original MGCN-FLC model.}\label{tab:table3}
    \centering
{\fontsize{6}{7}\selectfont
    \begin{tabular}{l|c|c|c}
\noalign{\hrule height 1pt}
\multirow{2}{*}{\textbf{Datasets/Methods}} & \multicolumn{3}{c}{\textbf{MGCN-FLC}} \\ \cline{2-4}
& \textbf{VA-based} & \textbf{VSF-based} & \textbf{IF-based (our)} \\
 \hline
\begin{tabular}{p{1.5cm}p{0.5cm}}BBCnews& \begin{tabular}{l}ACC \\ F1\end{tabular} \\\end{tabular}  &  \begin{tabular}{c}\underline{98.0}\\\underline{97.2}\end{tabular} & \begin{tabular}{c}96.6 \\95.5\end{tabular} &  \begin{tabular}{c}{\bf 98.4} \\{\bf97.9}\end{tabular}\\
\hline
\begin{tabular}{p{1.5cm}p{0.5cm}}Caltech101-7& \begin{tabular}{l}ACC \\ F1\end{tabular} \\\end{tabular}   & \begin{tabular}{c}\underline{79.0} \\\underline{55.2}\end{tabular} &  \begin{tabular}{c}69.8 \\40.2\end{tabular} &  \begin{tabular}{c}{\bf97.5} \\{\bf91.3}\end{tabular}\\
\hline
\begin{tabular}{p{1.5cm}p{0.5cm}}MNIST& \begin{tabular}{l}ACC \\ F1\end{tabular} \\\end{tabular}  & \begin{tabular}{c}\underline{95.6} \\\underline{93.5}\end{tabular} & \begin{tabular}{c}81.2\\78.5\end{tabular} &  \begin{tabular}{c}{\bf96.4 }\\{\bf95.8}\end{tabular}\\
 \hline
\begin{tabular}{p{1.5cm}p{0.5cm}}NUS-WIDE& \begin{tabular}{l}ACC \\ F1\end{tabular} \\\end{tabular}   & \begin{tabular}{c}\underline{79.0} \\\underline{78.9}\end{tabular} & \begin{tabular}{c}77.6 \\77.6\end{tabular} &  \begin{tabular}{c}{\bf99.0} \\{\bf99.0}\end{tabular}\\
 \hline
\begin{tabular}{p{1.5cm}p{0.5cm}}WebKB& \begin{tabular}{l}ACC \\ F1\end{tabular} \\\end{tabular}   & \begin{tabular}{c}\underline{90.2} \\\underline{65.4}\end{tabular} & \begin{tabular}{c}88.0 \\64.0\end{tabular} &  \begin{tabular}{c}{\bf91.8 }\\{\bf67.4}\end{tabular}\\
\hline
\begin{tabular}{p{1.5cm}p{0.5cm}}BBCsports& \begin{tabular}{l}ACC \\ F1\end{tabular} \\\end{tabular} & \begin{tabular}{c}\underline{96.9} \\\underline{96.7}\end{tabular} & \begin{tabular}{c}\underline{96.9} \\96.5\end{tabular} &  \begin{tabular}{c}{\bf97.3 }\\{\bf97.2}\end{tabular}\\
\hline
\begin{tabular}{p{1.5cm}p{0.5cm}}NGs& \begin{tabular}{l}ACC \\ F1\end{tabular} \\\end{tabular} & \begin{tabular}{c}\underline{97.4}\\\underline{97.4}\end{tabular} & \begin{tabular}{c}96.0\\96.0\end{tabular} &  \begin{tabular}{c}{\bf97.8 }\\{\bf97.8}\end{tabular}\\
\hline
\begin{tabular}{p{1.5cm}p{0.5cm}}ProteinFold
& \begin{tabular}{l}ACC \\ F1\end{tabular} \\\end{tabular}  & \begin{tabular}{c}27.8 \\15.6\end{tabular} & \begin{tabular}{c}\underline{42.5}\\\underline{28.5}\end{tabular} &  \begin{tabular}{c}{\bf60.0 }\\{\bf48.3}\end{tabular}\\
\hline
\begin{tabular}{p{1.5cm}p{0.5cm}}Reuters
& \begin{tabular}{l}ACC \\ F1\end{tabular} \\\end{tabular}  & \begin{tabular}{c}\underline{97.8} \\\underline{96.4}\end{tabular} & \begin{tabular}{c}91.0 \\87.7\end{tabular} &  \begin{tabular}{c}{\bf99.1 }\\{\bf98.9}\end{tabular}\\
\noalign{\hrule height 1pt}
    \end{tabular}
\label{table 6}}
\end{table}
To further validate the differences between the proposed interactive fusion module and existing interaction mechanisms, we introduced two variants of MGCN-FLC, replacing the interactive fusion module with view-level attention and view-shared underlying feature, respectively. The details of these two MGCN-FLC variants are as follows: (1) the variant based on view-level attention assigns learnable weights to each view; (2) the variant based on view-shared underlying feature integrates multi-view features into a shared feature matrix through a trainable fully connected network which is optimized by a reconstruction loss function. Table \ref{table 6} presents the classification performance of MGCN-FLC and its two variants on all datasets. Obviously, the proposed interactive fusion module outperforms the other two interaction mechanisms on all datasets. Compared to the MGCN-FLC variant based on view-level attention, the original MGCN-FLC based on interactive fusion module improves ACC by 18.5\%, 20\%, and 32.2\% on the Caltech101-7, NUS-WIDE, and ProteinFold datasets, respectively. These improvements stem from the fine-grained feature-level interactions between views conducted in the interactive fusion module, allowing for a deeper exploration of inter-view consistency. Compared to the MGCN-FLC variant based on view-shared underlying feature, the original MGCN-FLC based on interactive fusion module improves ACC by 27.7\%, 15.2\%, 21.4\%, 17.5\%, and 8.1\% on the Caltech101-7, MNIST, NUS-WIDE, ProteinFold, and Reuters datasets, respectively. These improvements stem from the more comprehensive pairwise inter-view consistency achieved by the interactive fusion module, rather than the narrower consistency among all views obtained by the view-shared underlying feature.

\subsection{Runtime comparison}
\begin{figure*}[ht]
\centering
\includegraphics[width=1.0\textwidth]{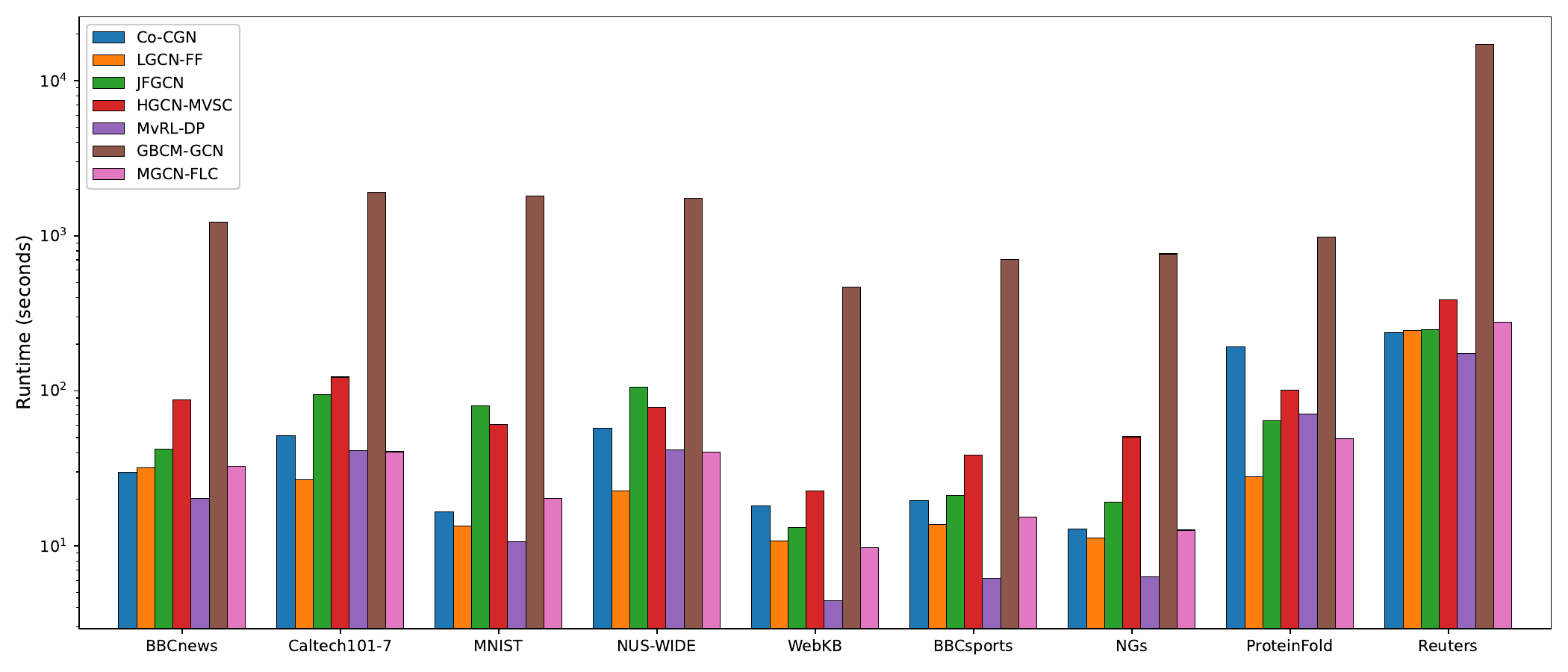}
\caption{Running time of MGCN-FLC and other GCN-based models on all datasets.}
\label{Runtime}
\end{figure*}

We compared the training time of MGCN-FLC and other GCN-based algorithms in Figure \ref{Runtime}. Due to the introduction of the GB algorithm \cite{DBLP:journals/tnn/ChengLXWHZ24} during topology construction, MGCN-FLC needs additional computational time; however, its overall training time remains within an acceptable range. Compared with GBCM-GCN \cite{DBLP:journals/eswa/WangYZXYX26}, which also employed the GB algorithm \cite{DBLP:journals/tnn/ChengLXWHZ24}, MGCN-FLC demonstrates higher training efficiency. Although MvRL-DP \cite{DBLP:journals/kbs/WangLWGW25} is the fastest among all algorithms, it suffers from severe performance degradation compared to our MGCN-FLC, particularly on the NUS-WIDE dataset, where its accuracy drops by 44.7\%. LGCN-FF \cite{DBLP:journals/inffus/ChenFYGPW23} ranks as the second fastest algorithm, it also encounters a significant accuracy gap when compared to MGCN-FLC, with particularly severe drops of 66.6\% and 34.2\% on the NUS-WIDE and Reuters datasets, respectively.

 \subsection{Experimental discussion}
 Our MGCN-FLC employs an unsupervised GB algorithm \cite{DBLP:journals/tnn/ChengLXWHZ24}, enabling it to achieve excellent performance on low-dimensional datasets, yet the model's performance may be constrained by the GB algorithm's \cite{DBLP:journals/tnn/ChengLXWHZ24} ability to handle high-dimensional sparse data. The results in Table \ref{table 4} show that on the high-dimensional sparse Reuters dataset, our MGCN-FLC outperforms the $MGCN-FLC_{tc}$ using the \textit{k}NN algorithm, achieving a 26.4\% improvement in ACC. This improvement is primarily attributed to the feature enhancement module and the interactive fusion module within MGCN-FLC, which effectively compensate for the GB algorithm's \cite{DBLP:journals/tnn/ChengLXWHZ24} limitations, enabling the model to outperform other algorithms.
 
 The complexity of the interactive fusion module in MGCN-FLC exhibits quadratic growth with respect to the number of views. However, the number of views is typically not high in practice, thus having limited impact on the overall complexity. To evaluate the efficiency of the interactive fusion module, we compare MGCN-FLC with its variant based on view-shared underlying feature. Table \ref{table 6} presents the classification performance of our MGCN-FLC and the variant on all datasets. For the ProteinFold dataset, the MGCN-FLC improves ACC by 17.5\% compared to the variant. Both the interactive fusion module and the view-shared underlying feature module employ a sparse encoder for feature alignment to facilitate subsequent feature fusion. The computational complexities of the interactive fusion module and the view-shared underlying feature module are $O(N  d^v \ d^1 + N  d^1  d + (V-1)  N d)$ and $O(N  d^v \ d^1 + N  d^1  d + N d^2)$, respectively. Taking the ProteinFold dataset ($N$=694, $V$=12, $d^v$=27) as an example, after incorporating its parameters and those of the sparse encoder ($d^1$=1024, $d$=256) into the complexity formula, the complexity of the interactive fusion module is approximately 82\% of that of the view-shared underlying feature module, while achieving a 17.5\% ACC improvement, demonstrating the effectiveness of the module.
\section{Conclusion}
This paper proposes the multi-view graph convolutional network with fully leveraging consistency via granular-ball-based topology construction, feature enhancement and interactive fusion (MGCN-FLC), which explores inter-node consistency, inter-feature consistency, and inter-view consistency. The method employs the unsupervised GB algorithm to construct the high-quality topologies that leverages both the intra-GB high similarity and the inter-GB global information. This topology effectively captures inter-node consistency. The feature enhancement module generates enriched feature representations through intra-view feature interactions, fully utilizing the inter-feature consistency. The interactive fusion module explicitly and fully exploits inter-view consistency to further improve the feature representations. Experiments demonstrate the effectiveness of the proposed MGCN-FLC model and validate the importance of inter-node consistency, inter-feature consistency, and inter-view consistency  in multi-view learning. In the future, given that the feature enhancement module uses full-range feature similarity, lower similarity may introduce low-relevance information into the enhanced features, affecting their quality. To address this, we will attempt to cluster the features of each view and compute the similarity between features only within the cluster, which may help mitigate the impact of low-relevance information.
\section*{Acknowledgment}
This work was supported by National Science Foundation (Grant number 62466063, Grant number 61936001).
\bibliographystyle{unsrt}
\bibliography{ref.bib}

@inproceedings{DBLP:conf/cvpr/FazlaliXRL22,
	author       = {Hamidreza Fazlali and
	Yixuan Xu and
	Yuan Ren and
	Bingbing Liu},
	title        = {A Versatile Multi-View Framework for LiDAR-based 3D Object Detection
	with Guidance from Panoptic Segmentation},
	booktitle    = {{IEEE/CVF} Conference on Computer Vision and Pattern Recognition,
	{CVPR} 2022, New Orleans, LA, USA, June 18-24, 2022},
	pages        = {17171--17180},
	publisher    = {{IEEE}},
	year         = {2022}
}

@article{DBLP:journals/inffus/LupionRQSO24,
	author       = {Marcos Lupi{\'{o}}n and
	Aurora Polo Rodr{\'{\i}}guez and
	Javier Medina Quero and
	Juan F. Sanjuan and
	Pilar M. Ortigosa},
	title        = {3D Human Pose Estimation from multi-view thermal vision sensors},
	journal      = {Inf. Fusion},
	volume       = {104},
	pages        = {102154},
	year         = {2024}
}

@article{DBLP:journals/nn/WeiHWYZL24,
	author       = {Xing Wei and
	Taizhang Hu and
	Di Wu and
	Fan Yang and
	Chong Zhao and
	Yang Lu},
	title        = {{ECCT:} Efficient Contrastive Clustering via Pseudo-Siamese Vision
	Transformer and Multi-view Augmentation},
	journal      = {Neural Networks},
	volume       = {180},
	pages        = {106684},
	year         = {2024}
}

@article{DBLP:journals/tmm/HuangZFW23,
	author       = {Sheng Huang and
	Yunhe Zhang and
	Lele Fu and
	Shiping Wang},
	title        = {Learnable Multi-View Matrix Factorization With Graph Embedding and
	Flexible Loss},
	journal      = {{IEEE} Trans. Multim.},
	volume       = {25},
	pages        = {3259--3272},
	year         = {2023}
}

@article{DBLP:journals/isci/ZhangDLZCW23,
	author       = {Wei Zhang and
	Zhaohong Deng and
	Qiongdan Lou and
	Te Zhang and
	Kup{-}Sze Choi and
	Shitong Wang},
	title        = {Takagi-Sugeno-Kang Fuzzy System Towards Label-scarce Incomplete Multi-View
	Data Classification},
	journal      = {Inf. Sci.},
	volume       = {647},
	pages        = {119466},
	year         = {2023}
}

@inproceedings{DBLP:conf/aaai/XuSGZ0G24,
	author       = {Cai Xu and
	Jiajun Si and
	Ziyu Guan and
	Wei Zhao and
	Yue Wu and
	Xiyue Gao},
	editor       = {Michael J. Wooldridge and
	Jennifer G. Dy and
	Sriraam Natarajan},
	title        = {Reliable Conflictive Multi-View Learning},
	booktitle    = {Thirty-Eighth {AAAI} Conference on Artificial Intelligence, {AAAI}
	2024, Thirty-Sixth Conference on Innovative Applications of Artificial
	Intelligence, {IAAI} 2024, Fourteenth Symposium on Educational Advances
	in Artificial Intelligence, {EAAI} 2014, February 20-27, 2024, Vancouver,
	Canada},
	pages        = {16129--16137},
	publisher    = {{AAAI} Press},
	year         = {2024}
}

@inproceedings{cheng2021multi,
  author={Jiafeng Cheng and Qianqian Wang and Zhiqiang Tao and Deyan Xie and Quanxue Gao},
  title={Multi-view attribute graph convolution networks for clustering},
  booktitle={Proceedings of the twenty-ninth international conference on international joint conferences on artificial intelligence},
  pages={2973--2979},
  year={2021}
}

@inproceedings{DBLP:conf/nips/WuZF23,
	author       = {Zhihao Wu and
	Zhao Zhang and
	Jicong Fan},
	editor       = {Alice Oh and
	Tristan Naumann and
	Amir Globerson and
	Kate Saenko and
	Moritz Hardt and
	Sergey Levine},
	title        = {Graph Convolutional Kernel Machine versus Graph Convolutional Networks},
	booktitle    = {Advances in Neural Information Processing Systems 36: Annual Conference
	on Neural Information Processing Systems 2023, NeurIPS 2023, New Orleans,
	LA, USA, December 10 - 16, 2023},
	year         = {2023}
}

@article{DBLP:journals/eswa/WangZYHCW22,
  author       = {Lijuan Wang and
                  Lin Zhang and
                  Ming Yin and
                  Zhifeng Hao and
                  Ruichu Cai and
                  Wen Wen},
  title        = {Double embedding-transfer-based multi-view spectral clustering},
  journal      = {Expert Syst. Appl.},
  volume       = {210},
  pages        = {118374},
  year         = {2022}
}

@article{DBLP:journals/nn/HuangZLYZZ25,
  author       = {Xin Huang and
                  Ranqiao Zhang and
                  Yuanyuan Li and
                  Fan Yang and
                  Zhiqin Zhu and
                  Zhihao Zhou},
  title        = {{MFC-ACL:} Multi-view fusion clustering with attentive contrastive
                  learning},
  journal      = {Neural Networks},
  volume       = {184},
  pages        = {107055},
  year         = {2025}
}

@article{wang2024graph,
  author={Zixiao Wang and Jicong Fan},
  title={Graph classification via reference distribution learning: theory and practice},
  journal={Advances in Neural Information Processing Systems},
  volume={37},
  pages={137698--137740},
  year={2024}
}

@article{yu2024sdhgcn,
  author={Bin Yu and Hengjie Xie and Jingxuan Chen and Mingjie Cai and Hamido Fujita and Weiping Ding},
  title={SDHGCN: A Heterogeneous Graph Convolutional Neural Network Combined With Shadowed Set},
  journal={IEEE Transactions on Fuzzy Systems},
  year={2024},
  publisher={IEEE}
}

@article{DBLP:journals/isci/ZhongLCSW24,
	author       = {Luying Zhong and
	Jielong Lu and
	Zhaoliang Chen and
	Na Song and
	Shiping Wang},
	title        = {Adaptive multi-channel contrastive graph convolutional network with
	graph and feature fusion},
	journal      = {Inf. Sci.},
	volume       = {658},
	pages        = {120012},
	year         = {2024}
}

@inproceedings{DBLP:conf/iclr/KipfW17,
	author       = {Thomas N. Kipf and
	Max Welling},
	title        = {Semi-Supervised Classification with Graph Convolutional Networks},
	booktitle    = {5th International Conference on Learning Representations, {ICLR} 2017,
	Toulon, France, April 24-26, 2017, Conference Track Proceedings},
	publisher    = {OpenReview.net},
	year         = {2017}
}

@inproceedings{DBLP:conf/kdd/SongZK22,
	author       = {Zixing Song and
	Yifei Zhang and
	Irwin King},
	editor       = {Aidong Zhang and
	Huzefa Rangwala},
	title        = {Towards an Optimal Asymmetric Graph Structure for Robust Semi-supervised
	Node Classification},
	booktitle    = {{KDD} '22: The 28th {ACM} {SIGKDD} Conference on Knowledge Discovery
	and Data Mining, Washington, DC, USA, August 14 - 18, 2022},
	pages        = {1656--1665},
	publisher    = {{ACM}},
	year         = {2022}
}

@inproceedings{DBLP:conf/kdd/YueLCB22,
	author       = {Qin Yue and
	Jiye Liang and
	Junbiao Cui and
	Liang Bai},
	editor       = {Aidong Zhang and
	Huzefa Rangwala},
	title        = {Dual Bidirectional Graph Convolutional Networks for Zero-shot Node
	Classification},
	booktitle    = {{KDD} '22: The 28th {ACM} {SIGKDD} Conference on Knowledge Discovery
	and Data Mining, Washington, DC, USA, August 14 - 18, 2022},
	pages        = {2408--2417},
	publisher    = {{ACM}},
	year         = {2022}
}

@article{DBLP:journals/tmm/LuWZCZW24,
	author       = {Jielong Lu and
	Zhihao Wu and
	Luying Zhong and
	Zhaoliang Chen and
	Hong Zhao and
	Shiping Wang},
	title        = {Generative Essential Graph Convolutional Network for Multi-View Semi-Supervised
	Classification},
	journal      = {{IEEE} Trans. Multim.},
	volume       = {26},
	pages        = {7987--7999},
	year         = {2024}
}

@article{DBLP:journals/tkdd/AtaFWSKL21,
	author       = {Sezin Kircali Ata and
	Yuan Fang and
	Min Wu and
	Jiaqi Shi and
	Chee Keong Kwoh and
	Xiaoli Li},
	title        = {Multi-View Collaborative Network Embedding},
	journal      = {{ACM} Trans. Knowl. Discov. Data},
	volume       = {15},
	number       = {3},
	pages        = {39:1--39:18},
	year         = {2021}
}

@article{DBLP:journals/corr/abs-2206-04216,
	author       = {Seongjun Yun and
	Seoyoon Kim and
	Junhyun Lee and
	Jaewoo Kang and
	Hyunwoo J. Kim},
	title        = {Neo-GNNs: Neighborhood Overlap-aware Graph Neural Networks for Link
	Prediction},
	journal      = {CoRR},
	volume       = {abs/2206.04216},
	year         = {2022}
}

@article{DBLP:journals/inffus/ChenFYGPW23,
	author       = {Zhaoliang Chen and
	Lele Fu and
	Jie Yao and
	Wenzhong Guo and
	Claudia Plant and
	Shiping Wang},
	title        = {Learnable graph convolutional network and feature fusion for multi-view
	learning},
	journal      = {Inf. Fusion},
	volume       = {95},
	pages        = {109--119},
	year         = {2023}
}

@article{DBLP:journals/eswa/PengDC25,
	author       = {Guowen Peng and
	Fadi Dornaika and
	Jinan Charafeddine},
	title        = {{CGCN-FMF:1D} convolutional neural network based feature fusion and
	multi graph fusion for semi-supervised learning},
	journal      = {Expert Syst. Appl.},
	volume       = {277},
	pages        = {127194},
	year         = {2025}
}

@article{DBLP:journals/kbs/WangLWGW25,
	author       = {Xuzheng Wang and
	Shiyang Lan and
	Zhihao Wu and
	Wenzhong Guo and
	Shiping Wang},
	title        = {Multi-view Representation Learning with Decoupled private and shared
	Propagation},
	journal      = {Knowl. Based Syst.},
	volume       = {310},
	pages        = {112956},
	year         = {2025}
}

@article{DBLP:journals/eswa/WangYZXYX26,
  author       = {Weijun Wang and
                  Xibei Yang and
                  Qinghua Zhang and
                  Shuyin Xia and
                  Jie Yang and
                  Taihua Xu},
  title        = {Capturing local and global information: Multi-view graph convolutional
                  network via granular-ball computing and collaborative matrix},
  journal      = {Expert Syst. Appl.},
  volume       = {296},
  pages        = {129057},
  year         = {2026}
}

@inproceedings{DBLP:conf/aaai/LiLW20a,
	author       = {Shu Li and
	Wen{-}Tao Li and
	Wei Wang},
	title        = {Co-GCN for Multi-View Semi-Supervised Learning},
	booktitle    = {The Thirty-Fourth {AAAI} Conference on Artificial Intelligence, {AAAI}
	2020, The Thirty-Second Innovative Applications of Artificial Intelligence
	Conference, {IAAI} 2020, The Tenth {AAAI} Symposium on Educational
	Advances in Artificial Intelligence, {EAAI} 2020, New York, NY, USA,
	February 7-12, 2020},
	pages        = {4691--4698},
	publisher    = {{AAAI} Press},
	year         = {2020}
}

@article{DBLP:journals/nn/ChenWCDW23,
	author       = {Yuhong Chen and
	Zhihao Wu and
	Zhaoliang Chen and
	Mianxiong Dong and
	Shiping Wang},
	title        = {Joint learning of feature and topology for multi-view graph convolutional
	network},
	journal      = {Neural Networks},
	volume       = {168},
	pages        = {161--170},
	year         = {2023}
}

@article{DBLP:journals/tcsv/WangCFWZ23,
  author       = {Yiming Wang and
                  Dongxia Chang and
                  Zhiqiang Fu and
                  Jie Wen and
                  Yao Zhao},
  title        = {Incomplete Multiview Clustering via Cross-View Relation Transfer},
  journal      = {{IEEE} Trans. Circuits Syst. Video Technol.},
  volume       = {33},
  number       = {1},
  pages        = {367--378},
  year         = {2023}
}

@inproceedings{DBLP:conf/aaai/YangYPYF23,
  author       = {Xiaocheng Yang and
                  Mingyu Yan and
                  Shirui Pan and
                  Xiaochun Ye and
                  Dongrui Fan},
  editor       = {Brian Williams and
                  Yiling Chen and
                  Jennifer Neville},
  title        = {Simple and Efficient Heterogeneous Graph Neural Network},
  booktitle    = {Thirty-Seventh {AAAI} Conference on Artificial Intelligence, {AAAI}
                  2023, Thirty-Fifth Conference on Innovative Applications of Artificial
                  Intelligence, {IAAI} 2023, Thirteenth Symposium on Educational Advances
                  in Artificial Intelligence, {EAAI} 2023, Washington, DC, USA, February
                  7-14, 2023},
  pages        = {10816--10824},
  publisher    = {{AAAI} Press},
  year         = {2023}
}

@article{DBLP:journals/tmm/WuLLCBW23,
  author       = {Zhihao Wu and
                  Xincan Lin and
                  Zhenghong Lin and
                  Zhaoliang Chen and
                  Yang Bai and
                  Shiping Wang},
  title        = {Interpretable Graph Convolutional Network for Multi-View Semi-Supervised
                  Learning},
  journal      = {{IEEE} Trans. Multim.},
  volume       = {25},
  pages        = {8593--8606},
  year         = {2023}
}

@article{DBLP:journals/nn/DornaikaBCX25,
  author       = {Fadi Dornaika and
                  Jingjun Bi and
                  Jinan Charafeddine and
                  H. Xiao},
  title        = {Semi-supervised learning for multi-view and non-graph data using Graph
                  Convolutional Networks},
  journal      = {Neural Networks},
  volume       = {185},
  pages        = {107218},
  year         = {2025}
}

@article{DBLP:journals/tnn/ChengLXWHZ24,
	author       = {Dongdong Cheng and
	Ya Li and
	Shuyin Xia and
	Guoyin Wang and
	Jinlong Huang and
	Sulan Zhang},
	title        = {A Fast Granular-Ball-Based Density Peaks Clustering Algorithm for
	Large-Scale Data},
	journal      = {{IEEE} Trans. Neural Networks Learn. Syst.},
	volume       = {35},
	number       = {12},
	pages        = {17202--17215},
	year         = {2024}
}

@article{DBLP:journals/tkde/BiDFWHZ24,
  author       = {Wendong Bi and
                  Lun Du and
                  Qiang Fu and
                  Yanlin Wang and
                  Shi Han and
                  Dongmei Zhang},
  title        = {Make Heterophilic Graphs Better Fit {GNN:} {A} Graph Rewiring Approach},
  journal      = {{IEEE} Trans. Knowl. Data Eng.},
  volume       = {36},
  number       = {12},
  pages        = {8744--8757},
  year         = {2024}
}

@inproceedings{DBLP:conf/www/DuSFMLHZ22,
  author       = {Lun Du and
                  Xiaozhou Shi and
                  Qiang Fu and
                  Xiaojun Ma and
                  Hengyu Liu and
                  Shi Han and
                  Dongmei Zhang},
  editor       = {Fr{\'{e}}d{\'{e}}rique Laforest and
                  Rapha{\"{e}}l Troncy and
                  Elena Simperl and
                  Deepak Agarwal and
                  Aristides Gionis and
                  Ivan Herman and
                  Lionel M{\'{e}}dini},
  title        = {{GBK-GNN:} Gated Bi-Kernel Graph Neural Networks for Modeling Both
                  Homophily and Heterophily},
  booktitle    = {{WWW} '22: The {ACM} Web Conference 2022, Virtual Event, Lyon, France,
                  April 25 - 29, 2022},
  pages        = {1550--1558},
  publisher    = {{ACM}},
  year         = {2022}
}

@article{DBLP:journals/tkde/ZhaoYLZGZ23,
	author       = {Zhongying Zhao and
	Zhan Yang and
	Chao Li and
	Qingtian Zeng and
	Weili Guan and
	MengChu Zhou},
	title        = {Dual Feature Interaction-Based Graph Convolutional Network},
	journal      = {{IEEE} Trans. Knowl. Data Eng.},
	volume       = {35},
	number       = {9},
	pages        = {9019--9030},
	year         = {2023}
}

@article{DBLP:journals/corr/abs-2003-02587,
  author       = {Fuli Feng and
                  Xiangnan He and
                  Hanwang Zhang and
                  Tat{-}Seng Chua},
  title        = {Cross-GCN: Enhancing Graph Convolutional Network with k-Order Feature
                  Interactions},
  journal      = {CoRR},
  volume       = {abs/2003.02587},
  year         = {2020}
}

@inproceedings{xu2020deep,
  title={Deep embedded complementary and interactive information for multi-view classification},
  author={Xu, Jinglin and Li, Wenbin and Liu, Xinwang and Zhang, Dingwen and Liu, Ji and Han, Junwei},
  booktitle={Proceedings of the AAAI conference on artificial intelligence},
  volume={34},
  number={04},
  pages={6494--6501},
  year={2020}
}

@article{DBLP:journals/nn/WangHWLCZ24,
	author       = {Shiping Wang and
	Sujia Huang and
	Zhihao Wu and
	Rui Liu and
	Yong Chen and
	Dell Zhang},
	title        = {Heterogeneous graph convolutional network for multi-view semi-supervised
	classification},
	journal      = {Neural Networks},
	volume       = {178},
	pages        = {106438},
	year         = {2024}
}

@article{DBLP:journals/air/WangLCWHZ25,
  author       = {Shiping Wang and
                  Jiacheng Li and
                  Yuhong Chen and
                  Zhihao Wu and
                  Aiping Huang and
                  Le Zhang},
  title        = {Multi-scale graph diffusion convolutional network for multi-view learning},
  journal      = {Artif. Intell. Rev.},
  volume       = {58},
  number       = {6},
  pages        = {184},
  year         = {2025}
}

@article{DBLP:journals/eswa/WuCZWG25,
	author       = {Yilin Wu and
	Zhaoliang Chen and
	Ying Zou and
	Shiping Wang and
	Wenzhong Guo},
	title        = {Multi-scale structure-guided graph generation for multi-view semi-supervised
	classification},
	journal      = {Expert Syst. Appl.},
	volume       = {263},
	pages        = {125677},
	year         = {2025}
}

@article{DBLP:journals/isci/XiaLDWYL19,
	author       = {Shuyin Xia and
	Yunsheng Liu and
	Xin Ding and
	Guoyin Wang and
	Hong Yu and
	Yuoguo Luo},
	title        = {Granular ball computing classifiers for efficient, scalable and robust
	learning},
	journal      = {Inf. Sci.},
	volume       = {483},
	pages        = {136--152},
	year         = {2019}
}

@article{DBLP:journals/tnn/XiaDWGG24,
	author       = {Shuyin Xia and
	Xiaochuan Dai and
	Guoyin Wang and
	Xinbo Gao and
	Elisabeth Giem},
	title        = {An Efficient and Adaptive Granular-Ball Generation Method in Classification
	Problem},
	journal      = {{IEEE} Trans. Neural Networks Learn. Syst.},
	volume       = {35},
	number       = {4},
	pages        = {5319--5331},
	year         = {2024}
}

@article{DBLP:journals/tetc/HeCGS24,
	author       = {Meixia He and
	Jianrui Chen and
	Maoguo Gong and
	Zhongshi Shao},
	title        = {{HDGCN:} Dual-Channel Graph Convolutional Network With Higher-Order
	Information for Robust Feature Learning},
	journal      = {{IEEE} Trans. Emerg. Top. Comput.},
	volume       = {12},
	number       = {1},
	pages        = {126--138},
	year         = {2024}
}

@inproceedings{DBLP:conf/kdd/0017ZB0SP20,
	author       = {Xiao Wang and
	Meiqi Zhu and
	Deyu Bo and
	Peng Cui and
	Chuan Shi and
	Jian Pei},
	editor       = {Rajesh Gupta and
	Yan Liu and
	Jiliang Tang and
	B. Aditya Prakash},
	title        = {{AM-GCN:} Adaptive Multi-channel Graph Convolutional Networks},
	booktitle    = {{KDD} '20: The 26th {ACM} {SIGKDD} Conference on Knowledge Discovery
	and Data Mining, Virtual Event, CA, USA, August 23-27, 2020},
	pages        = {1243--1253},
	publisher    = {{ACM}},
	year         = {2020}
}

@article{DBLP:journals/isci/WangLCL21,
	author       = {Jie Wang and
	Jianqing Liang and
	Junbiao Cui and
	Jiye Liang},
	title        = {Semi-supervised learning with mixed-order graph convolutional networks},
	journal      = {Inf. Sci.},
	volume       = {573},
	pages        = {171--181},
	year         = {2021}
}

@article{DBLP:journals/nn/LiuGTQSX21,
	author       = {Wenfeng Liu and
	Maoguo Gong and
	Zedong Tang and
	A. Kai Qin and
	Kai Sheng and
	Mingliang Xu},
	title        = {Locality preserving dense graph convolutional networks with graph
	context-aware node representations},
	journal      = {Neural Networks},
	volume       = {143},
	pages        = {108--120},
	year         = {2021}
}

@inproceedings{DBLP:conf/aaai/NieCL17,
	author       = {Feiping Nie and
	Guohao Cai and
	Xuelong Li},
	editor       = {Satinder Singh and
	Shaul Markovitch},
	title        = {Multi-View Clustering and Semi-Supervised Classification with Adaptive
	Neighbours},
	booktitle    = {Proceedings of the Thirty-First {AAAI} Conference on Artificial Intelligence,
	February 4-9, 2017, San Francisco, California, {USA}},
	pages        = {2408--2414},
	publisher    = {{AAAI} Press},
	year         = {2017}
}

@article{DBLP:journals/tip/TaoHNZY17,
	author       = {Hong Tao and
	Chenping Hou and
	Feiping Nie and
	Jubo Zhu and
	Dongyun Yi},
	title        = {Scalable Multi-View Semi-Supervised Classification via Adaptive Regression},
	journal      = {{IEEE} Trans. Image Process.},
	volume       = {26},
	number       = {9},
	pages        = {4283--4296},
	year         = {2017}
}

@article{DBLP:journals/pami/WangCDL22,
	author       = {Shiping Wang and
	Zhaoliang Chen and
	Shide Du and
	Zhouchen Lin},
	title        = {Learning Deep Sparse Regularizers With Applications to Multi-View
	Clustering and Semi-Supervised Classification},
	journal      = {{IEEE} Trans. Pattern Anal. Mach. Intell.},
	volume       = {44},
	number       = {9},
	pages        = {5042--5055},
	year         = {2022}
}
\end{document}